\documentclass[10pt,twocolumn,letterpaper]{article}

\usepackage{cvpr}
\usepackage{times}
\usepackage{epsfig}
\usepackage{graphicx}
\usepackage{amsmath}
\usepackage{amssymb}
\usepackage{subfigure}
\usepackage{multirow}
\usepackage{tabularx}
\usepackage{float}
\usepackage{algorithm}  
\usepackage{algorithmic}
\usepackage{color}
\usepackage{url}

\def\WIDTHFIVE {0.19\linewidth}
\def\WIDTHSIX {0.23\linewidth}
\def\WIDTHTHREE {0.315\linewidth}
\def\WIDTHTWO {0.45\linewidth}

\usepackage[pagebackref=true,breaklinks=true,letterpaper=true,colorlinks,bookmarks=false,linkcolor=red]{hyperref}

\cvprfinalcopy % *** Uncomment this line for the final submission

\def\cvprPaperID{1520} % *** Enter the CVPR Paper ID here

\ifcvprfinal\fi
\begin{document}

%%%%%%%%% TITLE
\title{Attentive Generative Adversarial Network for Raindrop Removal \\ from A Single Image}

\author{
	Rui~Qian$^1$,
	Robby T. Tan$^{2,3}$\thanks{R.T.Tan's work in this research is supported by the National Research Foundation, Prime Ministers Office, Singapore under its International Research Centre in Singapore Funding Initiative. His research is also supported in part by Yale-NUS College Start-Up Grant.}, 
	Wenhan~Yang$^1$,
	Jiajun~Su$^1$, 
	and~Jiaying~Liu$^1$\thanks{Corresponding author. This work was supported by National Natural Science Foundation of China under contract No. 61772043 and CCF-Tencent Open Research Fund. We also gratefully acknowledge the support of NVIDIA Corporation with the GPU for this research.}
	\\
	\small
	$^1$Institute of Computer Science and Technology, Peking University, Beijing, P.R. China
	\\
	\small
	$^2$ National University of Singapore,
	\small
	$^3$ Yale-NUS College	 
}

\maketitle
\thispagestyle{empty}

%%%%%%%%% ABSTRACT
\begin{abstract}
Raindrops adhered to a glass window or camera lens can severely hamper the visibility of a background scene and degrade an image considerably. In this paper, we address the problem by visually removing raindrops, and thus transforming a raindrop degraded image into a clean one. The problem is intractable, since first the regions occluded by raindrops are not given. Second, the information about the background scene of the occluded regions is completely lost for most part. To resolve the problem, we apply an attentive generative network using adversarial training. Our main idea is to inject visual attention into both the generative and discriminative networks. During the training, our visual attention  learns about raindrop regions and their surroundings. Hence, by injecting this information, the generative network will pay more attention to the raindrop regions and the surrounding structures, and the discriminative network will be able to assess the local consistency of the restored regions. This injection of visual attention to both generative and discriminative networks is the main contribution of this paper. Our experiments show the effectiveness of our approach, which outperforms the state of the art methods quantitatively and qualitatively.
\end{abstract}

%%%%%%%%% BODY TEXT

%%%%%%%%% INTRODUCTION
%-------------------------------------------------------------------------
\section{Introduction}

\begin{figure}
	\centering
	\subfigure{
	\includegraphics[width=\WIDTHTWO]{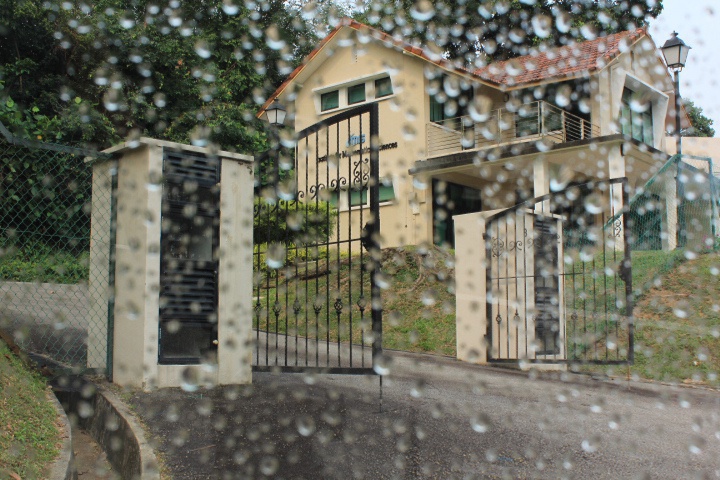}}
	\subfigure{
	\includegraphics[width=\WIDTHTWO]{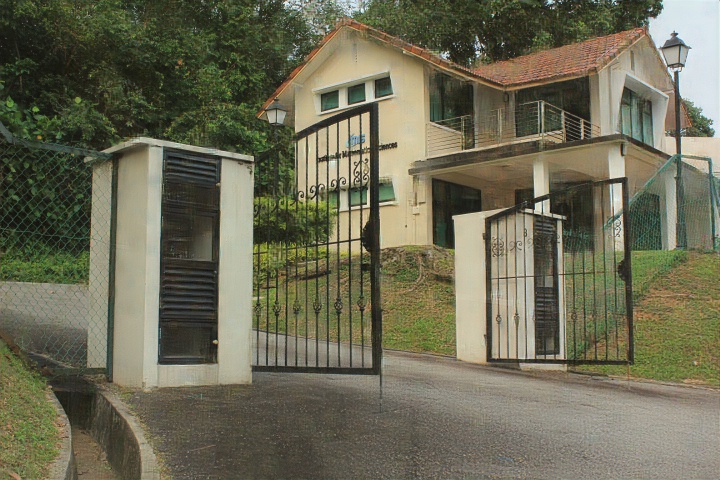}}\vfill\vspace{-2.5mm}
	\subfigure{
	\includegraphics[width=\WIDTHTWO]{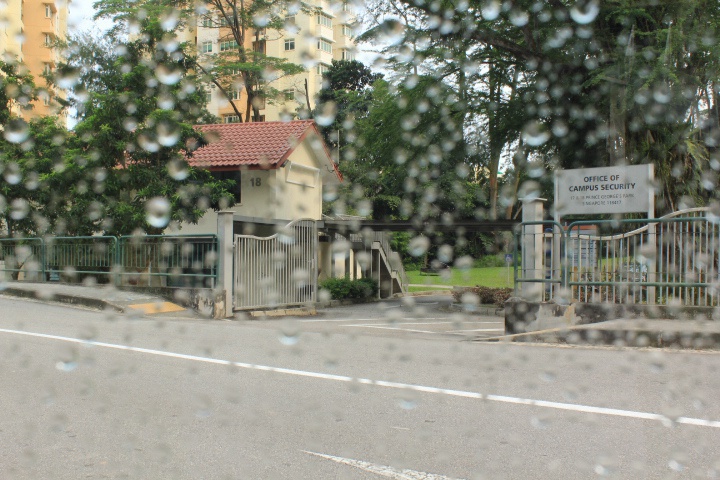}}
	\subfigure{
	\includegraphics[width=\WIDTHTWO]{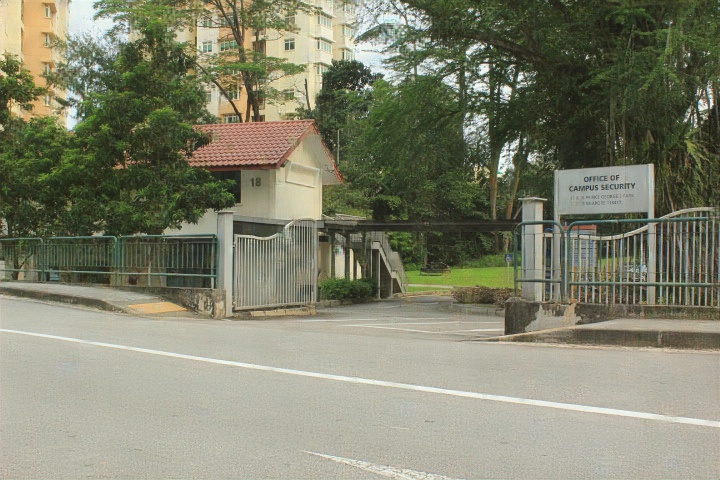}}\vfill\vspace{-2.5mm}
	\subfigure{
	\includegraphics[width=\WIDTHTWO]{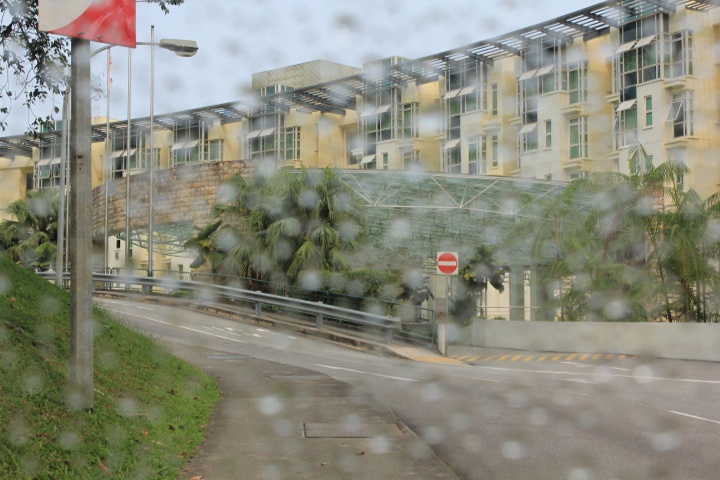}}
	\subfigure{
	\includegraphics[width=\WIDTHTWO]{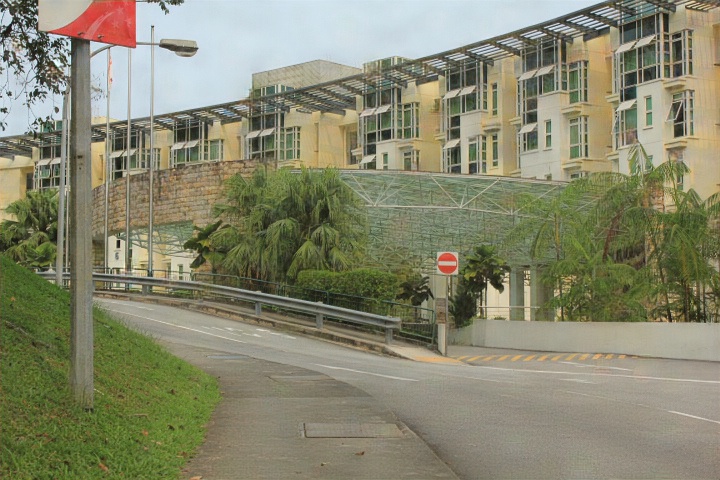}}
	\caption{Demonstration of our raindrop removal method. Left: input images degraded by raindrops. Right: our results, where most raindrops are removed and structural details are restored. Zooming-in the figure will provide a better look at the restoration quality.}
	\label{fig:trailer}
\end{figure}

Raindrops attached to a glass window, windscreen or lens can hamper the visibility of a background scene and degrade an image. Principally, the degradation occurs because raindrop regions contain different  imageries from those without raindrops.  Unlike non-raindrop regions,  raindrop regions are formed by rays of reflected light from a wider environment, due to the shape of raindrops, which is similar to that of a fish-eye lens. Moreover, in most cases, the focus of the camera is on the background scene, making the appearance of raindrops blur.

In this paper, we address this visibility degradation problem. Given an image impaired by raindrops, our goal is to remove the raindrops and produce a clean background as shown in Fig.~\ref{fig:trailer}. Our method is fully automatic. We consider that it will benefit image processing and computer vision applications, particularly for those suffering from raindrops, dirt, or similar artifacts.

A few methods have been proposed to tackle the raindrop detection and removal problems. Methods such as \cite{roser2009video,roser2010realistic,kurihata2005rainy} are dedicated to detecting raindrops but not removing them. Other methods are introduced to detect and remove raindrops using stereo \cite{tanaka2006removal}, video \cite{yamashita2009noises,you2016adherent}, or specifically designed optical shutter \cite{hara2009removal}, and thus are not applicable for a single input image taken by a normal camera. A method by Eigen et al. \cite{eigen2013restoring} has a similar setup to ours. It attempts to remove raindrops or dirt using a single image via deep learning method. However, it can only handle small raindrops, and produce blurry outputs \cite{you2016adherent}. In our experimental results (Sec. \ref{sec:results}), we will find that the method fails to handle relatively large and dense raindrops.

In contrast to \cite{eigen2013restoring}, we intend to deal with substantial presence of raindrops, like the ones shown in Fig.~\ref{fig:trailer}.  Generally, the raindrop-removal problem is intractable, since first the regions which are occluded by raindrops are not given. Second, the information about the background scene of the occluded regions is completely lost for most part. The problem gets worse when the raindrops are relatively large and distributed densely across the input image.
To resolve the problem, we use a generative adversarial network, where our generated outputs will be assessed by our discriminative network to ensure that our outputs look like real images. To deal with the complexity of the problem, our generative network first attempts to produce an attention map. This attention map is the most critical part of our network, since it will  guide the next process in the generative network to focus on raindrop regions.
This map is produced by a recurrent network consisting of deep residual networks (ResNets) \cite{he2016deep} combined with a convolutional LSTM~\cite{xingjian2015convolutional} and a few standard convolutional layers. We call this attentive-recurrent network. 

The second part of our generative network is an autoencoder, which takes both the input image and the  attention map as the input. To obtain wider contextual information, in the decoder side of the autoencoder, we apply multi-scale losses.  Each of these losses compares the difference between the output of the convolutional layers and the corresponding ground truth that has been downscaled accordingly. The input of the convolutional layers is the features from a decoder layer. Besides these losses, for the final output of the autoencoder, we apply a perceptual loss to obtain a more global similarity to the ground truth. This final output is also the output of our generative network.

Having obtained the generative image output, our discriminative network will check if it is real enough.  Like  in a few inpainting methods (e.g. \cite{iizuka2017globally,li2017generative}), our discriminative network validates the image both globally and locally. However, unlike the case of inpainting, in our problem and particularly in the testing stage, the target raindrop regions are not given. Thus, there is no information on the local regions that the discriminative network can focus on. To address this problem, we utilize our attention map to guide the discriminative network toward local target regions. 

Overall, besides introducing a novel method of raindrop removal, our other main contribution is the injection of the attention map into both generative and discriminative networks, which is novel and works effectively in removing raindrops, as shown in our experiments in Sec. \ref{sec:results}.  We will release our code and dataset.

The rest of the paper is organized as follows. Section \ref{sec:related_work} discusses the related work in the fields of raindrop detection and removal, and in the fields of the CNN-based image inpainting.  Section \ref{sec:image_formation} explains the raindrop model in an image, which is the basis of our method. Section \ref{sec:our_method} describes our method, which is based on the generative adversarial network. Section \ref{sec:dataset} discusses how we obtain our synthetic and real images used for training our network. Section \ref{sec:results} shows our evaluations quantitatively and qualitatively. Finally, Section \ref{sec:conclusion} concludes our paper.

%-------------------------------------------------------------------------
%%%%%%%%% RELATED WORK
\section{Related Work}
\label{sec:related_work}

There are a few papers dealing with bad weather visibility enhancement, which mostly tackle haze or fog (e.g. \cite{tan2008visibility,he2011single,ren2016single}), and rain streaks (e.g. \cite{garg2007vision,fu2017clearing,li2017single,yang2017deep}). Unfortunately, we cannot apply these methods directly to raindrop removal, since the image formation and the constraints of raindrops attached to a glass window  or lens are different from haze, fog, or rain streaks.

A number of methods have been proposed to detect raindrops. Kurihata et al.'s \cite{kurihata2005rainy} learns the shape of raindrops using PCA, and attempts to match a region in the test image, with those of the learned raindrops. However, since raindrops are transparent and have various shapes, it is unclear how large the number of raindrops needs to be learned, how to guarantee that PCA can model the various appearance of raindrops, and how to prevent other regions locally similar to raindrops to be detected as raindrops. Roser and Geiger's \cite{roser2009video} proposes a method that compares a synthetically generated raindrop with a patch that potentially has a raindrop. The synthetic raindrops are assumed to be a sphere section, and later  assumed to be inclined sphere sections \cite{roser2010realistic}. These assumptions might work in some cases, yet cannot be generalized to handle all raindrops, since raindrops can have various shapes and sizes.

Yamashita et al.'s \cite{yamashita2005removal} uses a stereo system to detect and remove raindrops. It detects raindrops by comparing  the disparities measured by the stereo with the distance between the stereo cameras and glass surface. It then removes raindrops by replacing the raindrop regions with the  textures of the corresponding image regions, assuming the other image does not have raindrops that occlude the same background scene. A similar method using an image sequence, instead of stereo, is proposed in Yamashita et al.'s \cite{yamashita2009noises}. Recently, You et al.'s \cite{you2016adherent} introduces a motion based method for detecting raindrops, and video completion to remove detected raindrops. While these methods work in removing raindrops  to some extent, they cannot be applied directly to a single image. 

Eigen et al.'s \cite{eigen2013restoring} tackles  single-image raindrop removal, which to our knowledge, is the only method  in the literature dedicated to the problem. The basic idea of the method is to train a convolutional neural network with pairs of raindrop-degraded images and the corresponding raindrop-free images. Its CNN consists of 3 layers, where each has 512 neurons. While the method works, particularly for relatively sparse and small droplets as well as dirt, it cannot produce  clean results for large and dense raindrops.  Moreover, the output images are somehow blur. We suspect that all these are due to the limited capacity of the network and the deficiency in providing enough constraints through its losses. Sec. \ref{sec:results} shows the comparison between our results with this method's.

In our method, we utilize a GAN \cite{goodfellow2014generative} as the backbone of our network, which is recently popular in dealing with the image inpainting or completion problem (e.g. \cite{iizuka2017globally,li2017generative}). Like in our method, \cite{iizuka2017globally} uses global and local assessment in its discriminative network. However, in contrast to our method, in the image inpainting, the target regions are given, so that the local assessment (whether local regions are sufficiently real) can be carried out. Hence, we cannot apply the existing image inpainting methods directly to our problem. Another similar architecture is Pix2Pix \cite{isola2016image}, which translates one image to another image.  It proposes a conditional GAN that not only learns the mapping from input image to output image, but also learns a loss function to the train the mapping. This method is a general mapping, and not proposed specifically to handle raindrop removal. In Sec. \ref{sec:results}, we will show some evaluations between our method and Pix2Pix.

%-------------------------------------------------------------------------
%%%%%%%%% IMAGE FORMATION
\section{Raindrop Image Formation}
\label{sec:image_formation}

We model a raindrop degraded image as the combination of a background image and effect of the raindrops:
\begin{equation}
\mathbf{I} = \mathbf(1 - \mathbf{M})\odot\mathbf{B} + \mathbf{R} 
\label{eq:model}
\end{equation}
where $\mathbf{I}$ is the colored input image and $\mathbf{M}$ is the binary mask. In the mask, $\mathbf{M}(\mathbf{x}) = 1$ means the pixel $\mathbf{x}$ is part of a raindrop region, and otherwise means it is part of background regions. $\mathbf{B}$ is the background image and $\mathbf{R}$ is the effect brought by the raindrops, representing the complex mixture of the background information and the light reflected by the environment  and passing through the raindrops adhered to a lens or windscreen. Operator $\odot$ means element-wise multiplication.

%In this image formation, we assume the lens or windscreen has uniform optical properties.

Raindrops are in fact transparent. However, due to their shapes and refractive index, a pixel in a raindrop region is not only influenced by one point in the real world but by the whole environment \cite{you2016adherent}, making most part of raindrops seem to have their own imagery different from the background scene. Moreover, since our camera is assumed to focus on the background scene, this imagery inside a raindrop region is mostly blur. Some parts of the raindrops, particularly at the periphery and transparent regions, convey some information about the background. We notice that the information can be revealed and used by our network.

Based on the model (Eq.~(\ref{eq:model})), our goal is to obtain the background image $\mathbf{B}$ from a given input $\mathbf{I}$. To accomplish this, we create an attention map guided by the binary mask $\mathbf{M}$. 
Note that, for our training data, as shown in Fig. \ref{fig:dataset_sample}, to obtain the mask we simply subtract the image degraded by raindrops $\mathbf{I}$ with its corresponding clean image $\mathbf{B}$. We use a threshold to determine whether a pixel is part of a raindrop region. In practice, we set the threshold to 30 for all images in our training dataset. This simple thresholding is sufficient for our purpose of generating the attention map.

%-------------------------------------------------------------------------

\begin{figure*}[h]
	\begin{center}
		\includegraphics[width=1\textwidth]{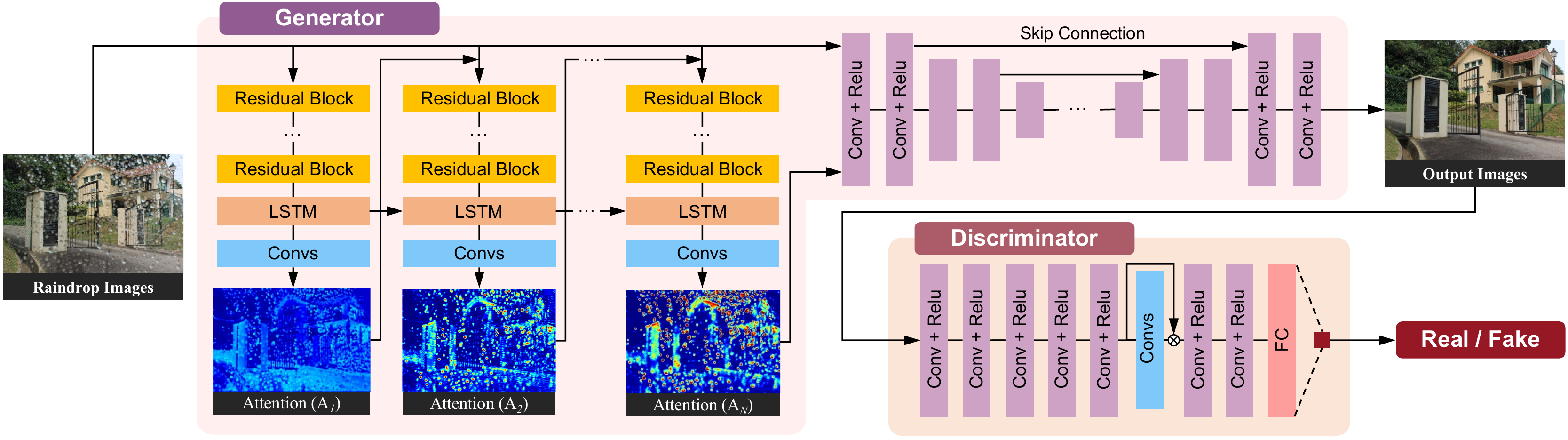}
	\end{center}
	\caption{The architecture of our proposed attentive GAN.The generator  consists of an attentive-recurrent network and a contextual autoencoder with skip connections. The discriminator is formed by a series of convolution layers and guided by the attention map. Best viewed in color.}
	\label{fig:network}
\end{figure*}

%%%%%%%%% ATTENTIVE GAN
\section{Raindrop Removal using Attentive GAN}
\label{sec:our_method}

Fig.~\ref{fig:network} shows the overall architecture of our proposed network. Following the  idea of  generative adversarial networks \cite{goodfellow2014generative}, there are two main parts in our network: the generative and discriminative networks.  Given an input image degraded by raindrops, our generative network attempts to produce an image as real as possible and free from raindrops. The discriminative network will validate whether the image produced by the generative network looks real. 

Our generative adversarial loss can be expressed as:
\begin{equation}
\begin{split}
\min \limits_{G} \max \limits_{D}  & \quad \mathbb{E}_{\mathbf{R} \sim p_{clean}} [log(D(\mathbf{R}))] \\ +
 & \quad \mathbb{E}_{\mathbf{I} \sim p_{raindrop}} [log(1 - D(G(\mathbf{I})))]
\end{split}
\end{equation}
where $G$ represents the generative network, and $D$ represents the discriminative network. $\mathbf{I}$ is a sample drawn from our pool of images degraded by raindrops, which is the input of our generative network. $\mathbf{R}$ is a sample from a pool of clean natural images.

\subsection{Generative Network}

As shown in Fig.~\ref{fig:network}, our generative network consists of two sub-networks: an attentive-recurrent network and a contextual autoencoder. The purpose of the attentive-recurrent network is to find regions in the input image that need to get attention. These regions are mainly the raindrop regions and their surrounding structures that are necessary for the contextual autoencoder to focus on, so that it can generate better local image restoration, and  for the discriminative network to focus the assessment on.

\paragraph{Attentive-Recurrent Network.}
Visual attention models have been applied to localizing targeted regions in an image to capture features of the regions.  The idea has been utilized for visual recognition and classification (e.g.~\cite{zhao2016diversified,mnih2014recurrent,gregor2015draw}). In a similar way, we consider visual attention to be important for generating raindrop-free background images, since it allows the network to know where the removal/restoration should be focused on. As shown in our architecture in Fig.~\ref{fig:network}, we employ a recurrent network to generate our visual attention.
Each block (of each time step) in our recurrent network comprises of five layers of ResNet \cite{he2016deep} that help extract features from the input image and the mask of the previous block, a convolutional LSTM unit \cite{xingjian2015convolutional}  and convolutional layers for generating the 2D attention maps.

\begin{figure}
	\centering
	\subfigure[Input]{
		\label{Fig.sub.1_e1}
		\includegraphics[width=0.15\textwidth]{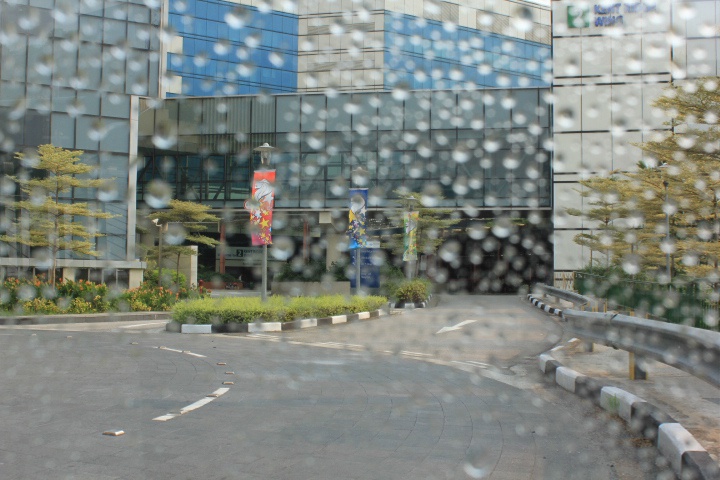}}
	\subfigure[epoch = 3]{
		\label{Fig.sub.2_e2}
		\includegraphics[width=0.15\textwidth]{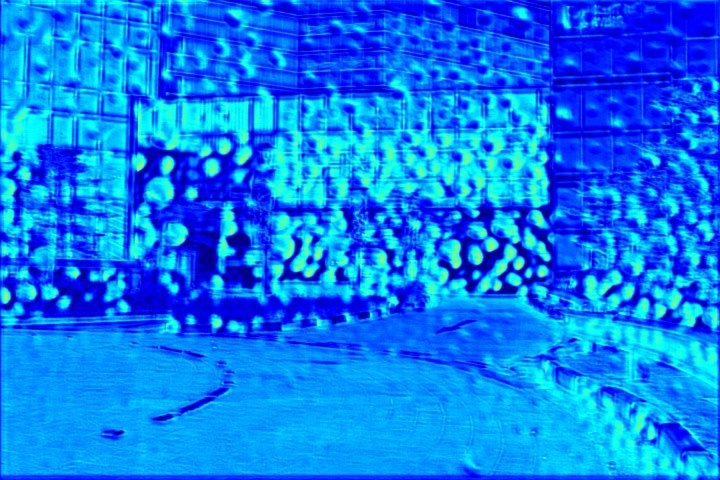}}
	\subfigure[epoch = 6]{
		\label{Fig.sub.3_e3}
		\includegraphics[width=0.15\textwidth]{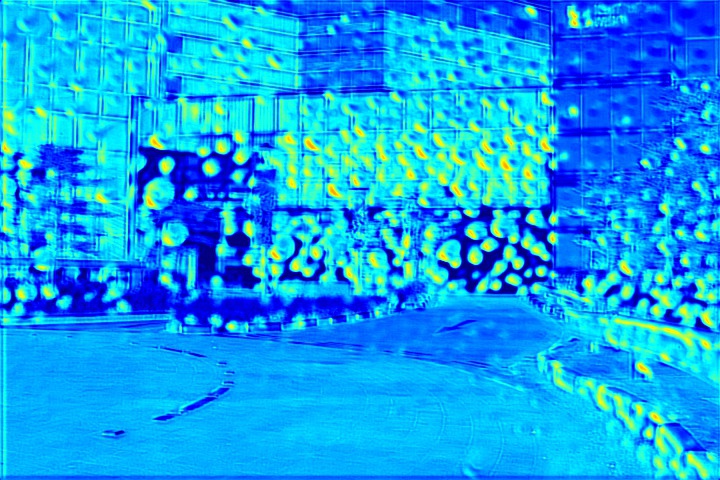}}
	\subfigure[epoch = 9]{
		\label{Fig.sub.4_e4}
		\includegraphics[width=0.15\textwidth]{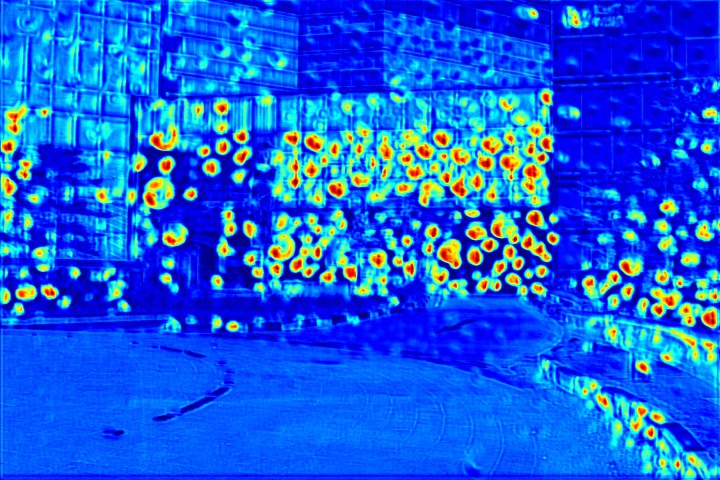}}
	\subfigure[epoch = 12]{
		\label{Fig.sub.5_e5}
		\includegraphics[width=0.15\textwidth]{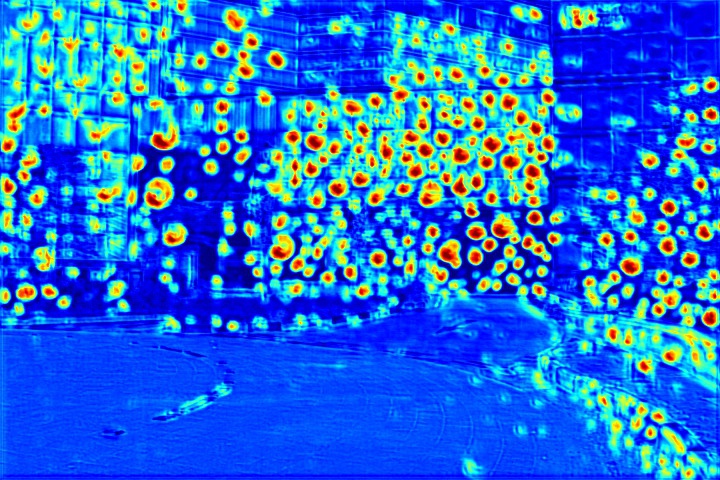}}
	\subfigure[epoch = 15]{
		\label{Fig.sub.6_e6}
		\includegraphics[width=0.15\textwidth]{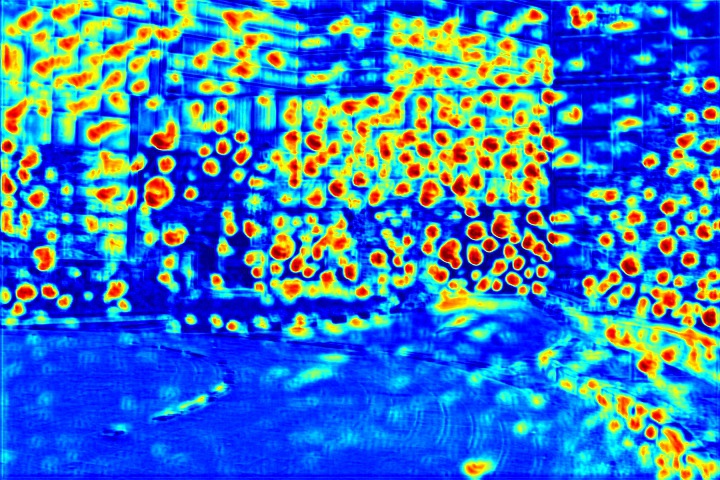}}
	\caption{Visualization of the attention map learning process. This visualization is for the final attention map, $\mathbf{A}_N$. Our attentive-recurrent network shows a greater focus on raindrop regions and the relevant structures during the training process.}
	\label{fig:attention_map}
\end{figure}

Our attention map, which is learned at each time step, is a matrix ranging from 0 to 1, where the greater the value, the greater attention it suggests, as shown in the visualization in Fig.~\ref{fig:attention_map}. Unlike the binary mask, $\mathbf{M}$, the attention map is a  non-binary map, and represents the increasing attention from non-raindrop regions to raindrop regions, and the values vary even inside raindrop regions. This increasing attention makes sense to have, since the surrounding regions of raindrops also needs the attention, and the transparency of a raindrop area in fact varies (some parts do not totally occlude the background, and thus convey some background information).

Our convolution LSTM unit consists of an input gate $\mathbf{i}_t$, a forget gate $\mathbf{f}_t$, an output gate $\mathbf{o}_t$ as well as a cell state $\mathbf{C}_t$. The interaction between states and gates along time dimension is defined as follows:
\begin{equation}
\begin{aligned}
\label{equ:LSTM}
  \mathbf{i}_t & =  \sigma(\mathbf{W}_{xi} * \mathbf{X}_t + \mathbf{W}_{hi} * \mathbf{H}_{t-1} + \mathbf{W}_{ci} \odot \mathbf{C}_{t-1} + \mathbf{b}_i)  \\
  \mathbf{f}_t & =  \sigma(\mathbf{W}_{xf} * \mathbf{X}_t + \mathbf{W}_{hf} * \mathbf{H}_{t-1} + \mathbf{W}_{cf} \odot \mathbf{C}_{t-1} + \mathbf{b}_f) \\
  \mathbf{C}_t & = \mathbf{f}_t \odot \mathbf{C}_{t-1} + \mathbf{i}_t \odot \tanh(\mathbf{W}_{xc}*\mathbf{X}_t + \mathbf{W}_{hc}*\mathbf{H}_{t-1} + \mathbf{b}_c) \\
  \mathbf{o}_t & =  \sigma(\mathbf{W}_{xo}*\mathbf{X}_t + \mathbf{W}_{ho}* \mathbf{H}_{t-1} + \mathbf{W}_{co} \odot\mathbf{C}_t + \mathbf{b}_o) \\
  \mathbf{H}_t & =  \mathbf{o}_t \odot \tanh(\mathbf{C}_t)\\
\end{aligned}
\end{equation}
where $\mathbf{X}_t$ is the features generated by ResNet. $\mathbf{C}_t$ encodes the cell state that will be fed to the next LSTM.  $\mathbf{H}_t$ represents the output features of the LSTM unit. Operator $*$ represents the convolution operation.
The LSTM's output feature is then fed into the convolutional layers, which generate  a 2D attention map. In the training process, we initialize the values of the attention map to 0.5. In each time step, we concatenate the current attention map  with the input image and then feed them into the next block of our recurrent network.

In training the generative network, we use pairs of images with and without raindrops that contain exactly the same background scene. The loss function in each recurrent block is defined as the mean squared error (MSE) between the output attention map at time step $t$, or  $\mathbf{A}_t$, and the binary mask, $\mathbf{M}$. We  apply this process $N$ time steps. The earlier attention maps have smaller values and get larger when approaching the $N^{th}$ time step indicating the increase in confidence. The loss function is expressed as:
\begin{equation}
\mathcal{L}_{ATT}(\{\mathbf{A}\}, \mathbf{M}) = \sum\limits^{N}_{t = 1} \theta^{N-t}\mathcal{L}_{MSE}(\mathbf{A}_t, \mathbf{M})
\end{equation}
where $\mathbf{A}_t$ is the attention map produced by the attentive-recurrent network at time step $t$. $\mathbf{A}_t = ATT_t(\mathbf{F}_{t-1}, \mathbf{H}_{t-1}, \mathbf{C}_{t-1})$, with $\mathbf{F}_{t-1}$  is  the concatenation of the input image and the attention map from the previous time step. When $t=1$, $\mathbf{F}_{t-1}$ is the input image concatenated with an initial attention map with values of 0.5.
 Function $ATT_t$ represents the attentive-recurrent network at time step $t$.
We set $N$ to 4 and $\theta$ to 0.8. We expect a higher $N$ will produce a better attention map, but it also requires larger memory. 

\begin{figure*}[!htp]
	\centering
	\subfigure[Input]{
	\includegraphics[width=\WIDTHFIVE]{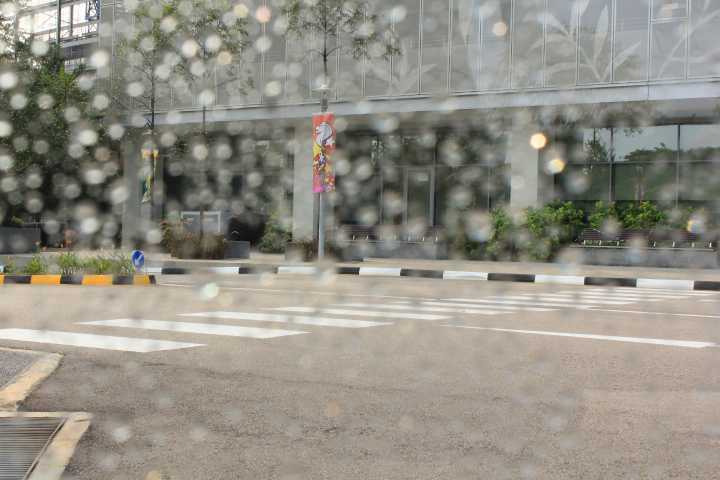}}
	\subfigure[Time step = 1]{
	\includegraphics[width=\WIDTHFIVE]{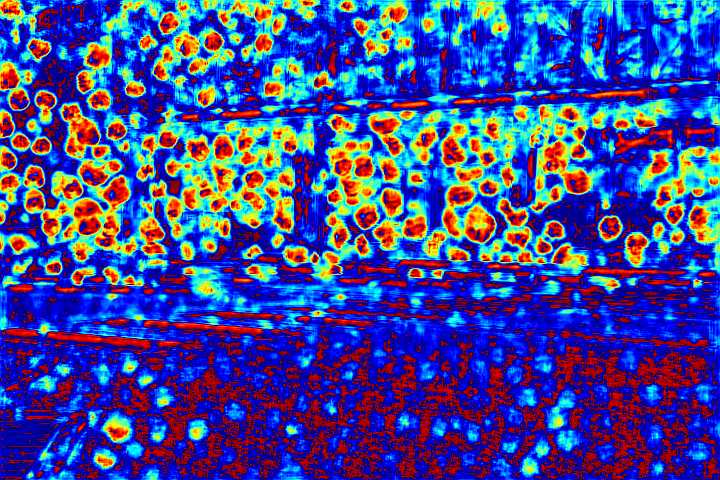}}
	\subfigure[Time step = 2]{
	\includegraphics[width=\WIDTHFIVE]{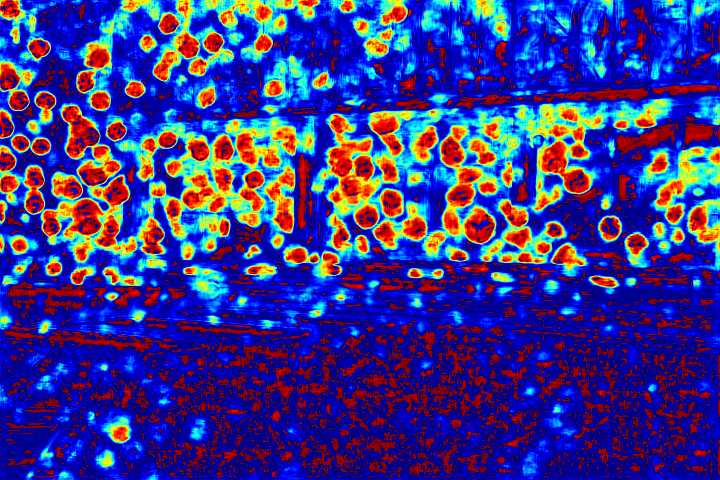}}
	\subfigure[Time step = 3]{
	\includegraphics[width=\WIDTHFIVE]{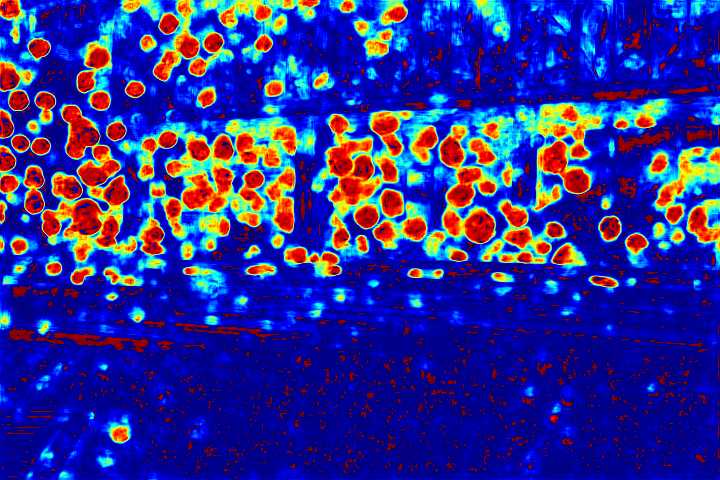}}
	\subfigure[Time step = 4]{
	\includegraphics[width=\WIDTHFIVE]{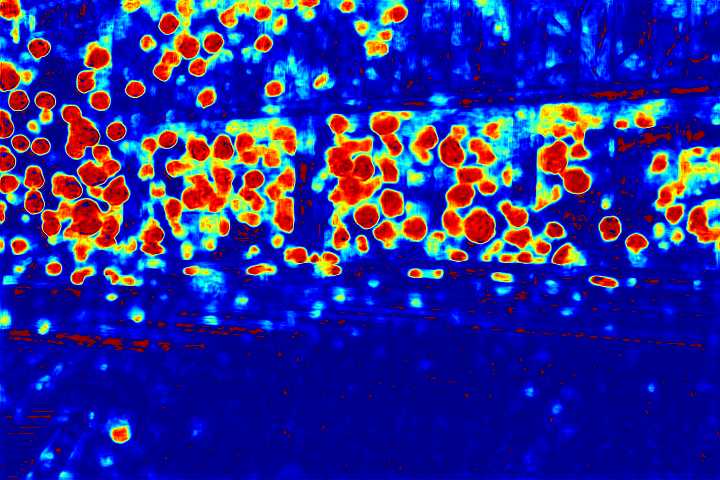}}
	\caption{Visualization of the attention map generated by our novel attentive-recurrent network. With the increasing of time step, our network focuses more and more on the raindrop regions and relevant structures. 
	}
	\label{fig:atten_step}
\end{figure*}

Fig.~\ref{fig:attention_map} shows some examples of attention maps generated by our network in the training procedure. As can be seen, our network attempts to find not only the raindrop regions but also some structures surrounding the regions. And Fig.~\ref{fig:atten_step} shows the effect of the attentive-recurrent network in the testing stage. With the increasing of time step, our network focuses more and more on the raindrop regions and relevant structures.

\begin{figure}
	\begin{center}
		\includegraphics[width=1\linewidth]{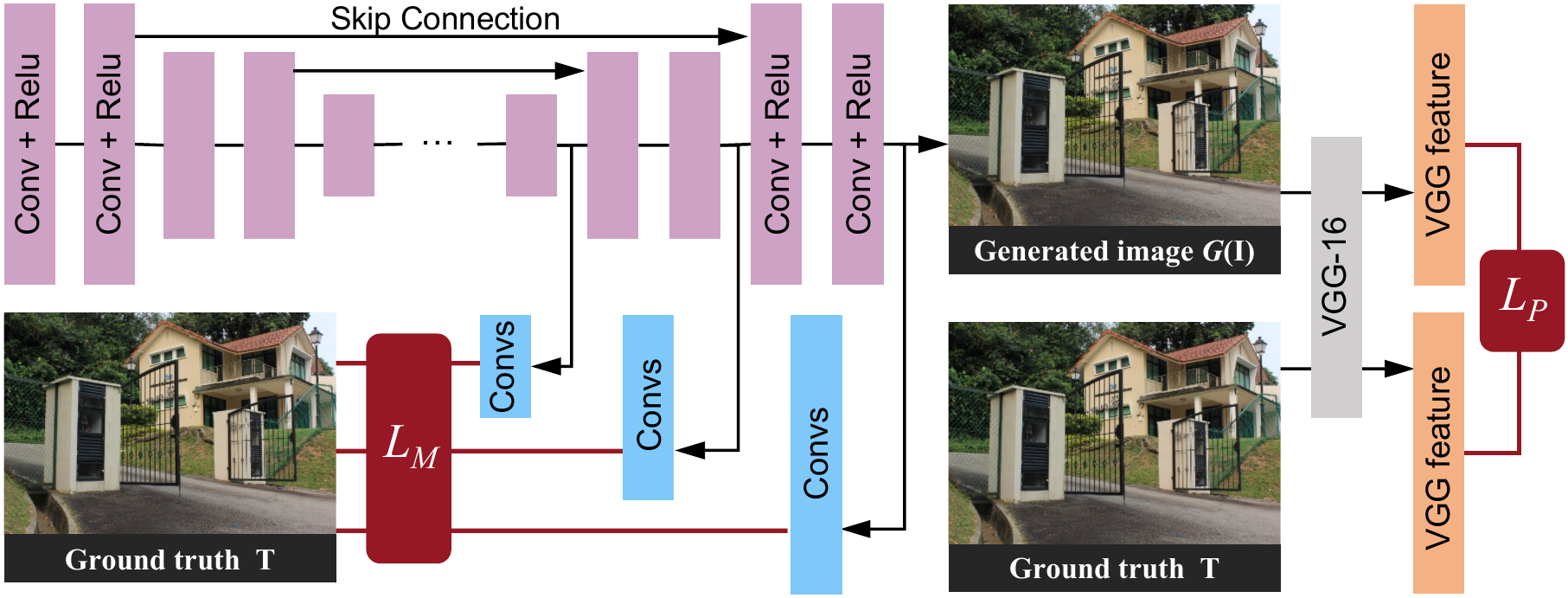}
	\end{center}
	\caption{The architecture of our contextual autoencoder. Multi-scale loss and perceptual loss are used to help train the autoencoder.}
	\label{fig_4}
\end{figure}

\paragraph{Contextual Autoencoder.}
The purpose of our contextual autoencoder is to generate an image that is free from raindrops. The input of the autoencoder is the concatenation of the input image and the final attention map from the attentive-recurrent network. Our deep autoencoder has 16 conv-relu blocks, and skip connections are added to prevent  blurred outputs. Fig.~\ref{fig_4} illustrates the architecture of our contextual autoencoder.

As shown in the figure, there are two loss functions in our autoencoder:  multi-scale losses and perceptual loss. For the multi-scale losses, we extract features from different decoder layers  to form outputs in different sizes. By adopting this, we intend to capture more contextual information from different scales. This is also the reason why we call it contextual autoencoder. 

We define the loss function as:
\begin{equation}
\mathcal{L}_M(\{\mathbf{S}\}, \{\mathbf{T}\}) = \sum\limits^M_{i = 1}\lambda_i \mathcal{L}_{MSE}(\mathbf{S}_i, \mathbf{T}_{i})
\end{equation}
where $\mathbf{S}_i$ indicates the $i$th output extracted from the decoder layers, and $\mathbf{T}_{i}$ indicates the ground truth that has the same scale as that of $\mathbf{S}_i$. ${\{\lambda_i\}}_{i=1}^M$ are the weights for different scales. We put more weight at the larger scale. 
To be more specific, the outputs of the last 1$^{st}$, 3$^{rd}$ and 5$^{th}$ layers are used, whose sizes are $\frac{1}{4}$, $\frac{1}{2}$ and $1$  of the  original size, respectively. Smaller layers are not used since the information is insignificant. We set $\lambda$'s to 0.6, 0.8, 1.0.

Aside from the multi-scale losses, which are based on a pixel-by-pixel operation, we also add a perceptual loss \cite{johnson2016perceptual} that measures the global discrepancy between the features of the autoencoder's output and those of the corresponding ground-truth clean image. These features can be   extracted from a well-trained CNN, \eg VGG16 pretrained on ImageNet dataset. Our perceptual loss function is expressed as:
\begin{equation}
\mathcal{L}_P(\mathbf{O},\mathbf{T}) = \mathcal{L}_{MSE}(VGG(\mathbf{O}), VGG(\mathbf{T}))
\end{equation}	
where $VGG$ is a pretrained CNN, and produces features from a given input image. 
$\mathbf{O}$ is the output image of the autoencoder or, in fact, of the whole generative network: $\mathbf{O} = G(\mathbf{I})$.  $\mathbf{T}$ is the ground-truth image that is free from raindrops.

Overall, the loss of our generative can be written as:
\begin{equation}
\label{eq_g}
\begin{split}
\mathcal{L}_G =  & \quad {{10}^{-2}} \mathcal{L}_{GAN}(\mathbf{O}) + \mathcal{L}_{ATT}(\{\mathbf{A}\},\mathbf{M})\\
& + \mathcal{L}_M(\{\mathbf{S}\}, \{\mathbf{T}\}) + \mathcal{L}_P(\mathbf{O}, \mathbf{T})
\end{split}
\end{equation}
where $\mathcal{L}_{GAN}(\mathbf{O}) = log(1 - D(\mathbf{O}))$.

\subsection{Discriminative Network}
To differentiate fake images from real ones, a few GAN-based methods adopt global and local image-content consistency in the discriminative part (e.g. \cite{iizuka2017globally,li2017generative}) . The global discriminator looks at the whole image to check if there is any inconsistency, while the local discriminator looks at small specific regions. The strategy of a local discriminator is particularly useful if we know the regions that are likely to be fake (like in the case of image inpainting, where the regions to be restored  are given). Unfortunately, in our problem, particularly in our testing stage, we do not know where the regions degraded by raindrops and the information is not given. Hence, the local discriminator must try to find those regions by itself.

To resolve this problem, our idea is to use an attentive discriminator. For this, we employ the attention map generated by our attentive-recurrent network. Specifically, we extract the features from the interior layers of the discriminator, and feed them to a CNN. We define a loss function based on the CNN's output and the attention map. Moreover, we use the CNN's output and  multiply it with the original features from the discriminative network, before feeding them into the next layers.  Our underlying idea of doing this is to guide our discriminator to focus on regions indicated by the attention map. Finally, at the end layer we use a fully connected layer to decide whether the input image is fake or real. The right part of Fig.~\ref{fig:network} illustrates our discriminative architecture. 

The whole loss function of the discriminator can be expressed as:
\begin{equation}
\label{eq_dis}
\begin{split}
\mathcal{L}_{D}(\mathbf{O},\mathbf{R}, \mathbf{A}_N) = &-log(D(\mathbf{R})) -log(1 - D(\mathbf{O})) \\
&+ \gamma \mathcal{L}_{map}(\mathbf{O}, \mathbf{R},  \mathbf{A}_N)
\end{split}
\end{equation}
where $\mathcal{L}_{map}$ is the loss between the features extracted from interior layers of the discriminator and the final attention map:
\begin{equation}
\begin{split}
\mathcal{L}_{map}(\mathbf{O},\mathbf{R}, \mathbf{A}_N) = &\mathcal{L}_{MSE}(D_{map}(\mathbf{O}), \mathbf{A}_N) \\
&+\mathcal{L}_{MSE}(D_{map}(\mathbf{R}), \mathbf{0})
\label{eq:L_map}
\end{split}
\end{equation}
where $D_{map}$ represents the process of producing a 2D map by the discriminative network. $\gamma$ is set to 0.05. $\mathbf{R}$ is a sample image drawn from a pool of real and clean images. $\mathbf{0}$ represents a map containing only 0 values. Thus, the second term of Eq.~(\ref{eq:L_map}) implies  that for $\mathbf{R}$, there is no specific region necessary to focus on. 

Our discriminative network contains 7 convolution layers with the kernel of (3, 3), a fully connected layer of 1024 and a single neuron with a sigmoid activation function. We extract the features from the last third convolution layers and multiply back in element-wise. 

%%%%%%%%% DATASET
\section{Raindrop Dataset}
\label{sec:dataset}
Similar to current deep learning methods, our method requires relatively a large amount of  data with groundtruths for training. However, since there is no such dataset for raindrops attached to a glass window or lens, we create our own. 
For our case, we need a set of image pairs, where each pair contains exactly the same background scene, yet one is degraded by raindrops and the other one is free from raindrops. To obtain this, we use two pieces of exactly the same glass: one sprayed with water, and the other is left clean. Using two pieces of glass allows us to avoid misalignment, as glass has a refractive index that is different from air, and thus refracts light rays. In general, we also need to manage any other causes of misalignment, such as camera motion, when taking the two images; and, ensure that the atmospheric conditions (e.g., sunlight, clouds, etc.) as well as  the background objects to be static during the acquisition process.

\begin{figure}
	\centering
	\subfigure{
	\includegraphics[width=\WIDTHSIX]{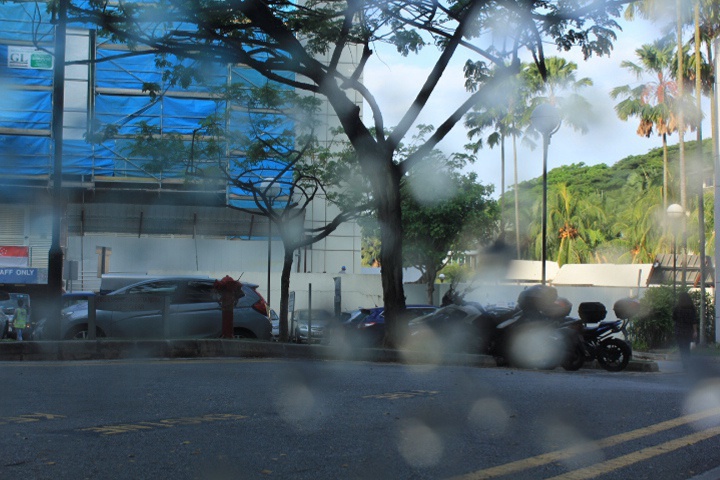}}
	\subfigure{
	\includegraphics[width=\WIDTHSIX]{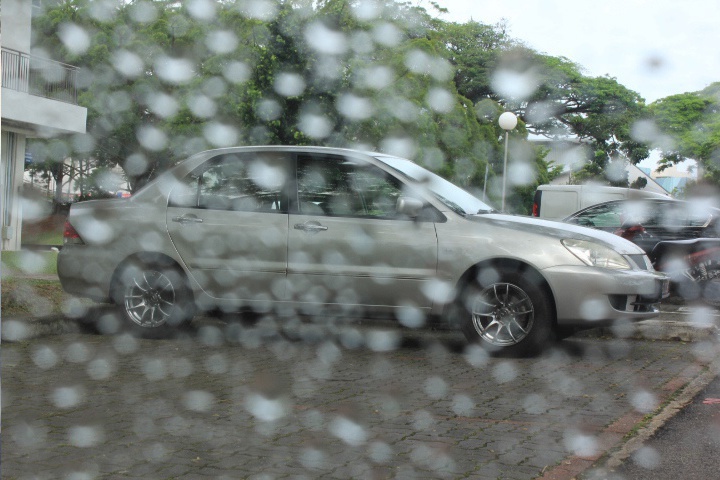}}
	\subfigure{
	\includegraphics[width=\WIDTHSIX]{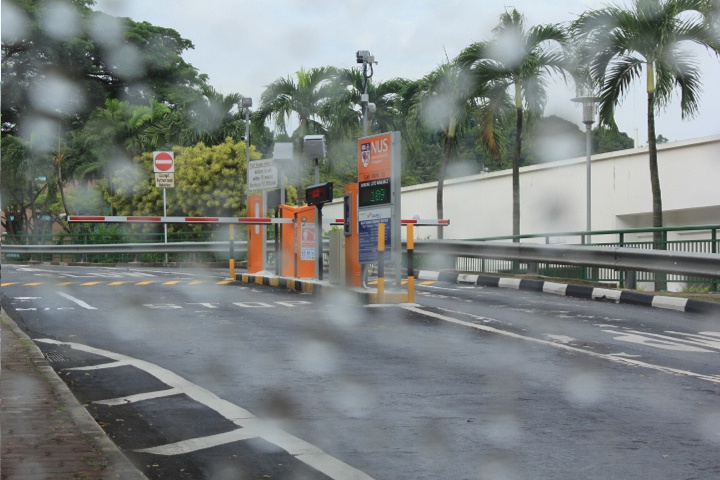}}
	\subfigure{
	\includegraphics[width=\WIDTHSIX]{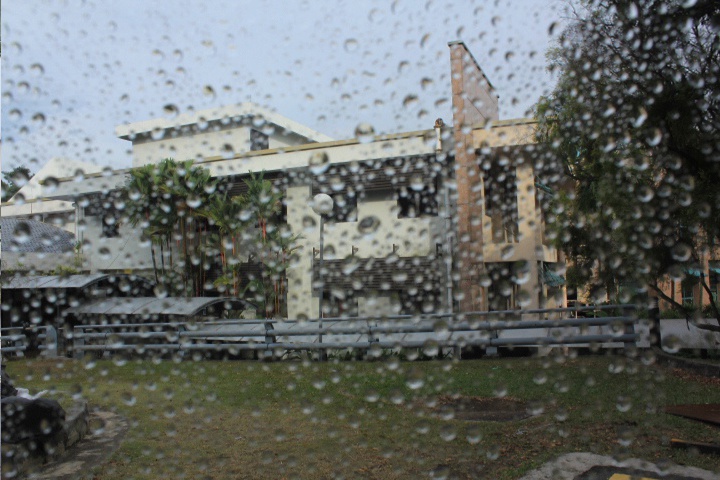}}\vfill\vspace{-2.5mm}
	\subfigure{
	\includegraphics[width=\WIDTHSIX]{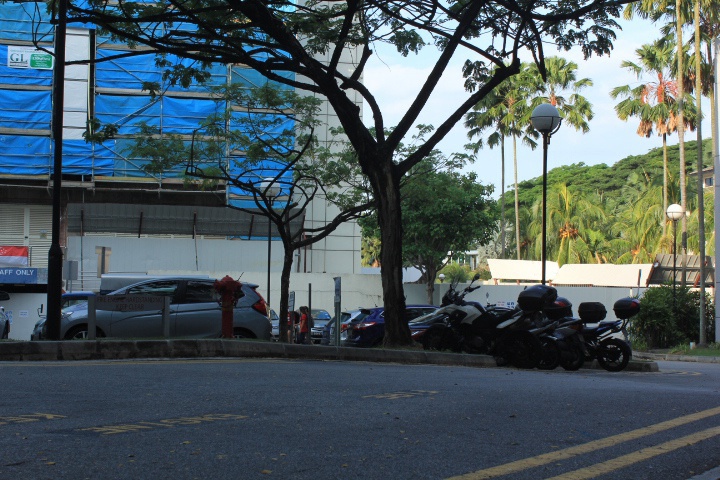}}
	\subfigure{
	\includegraphics[width=\WIDTHSIX]{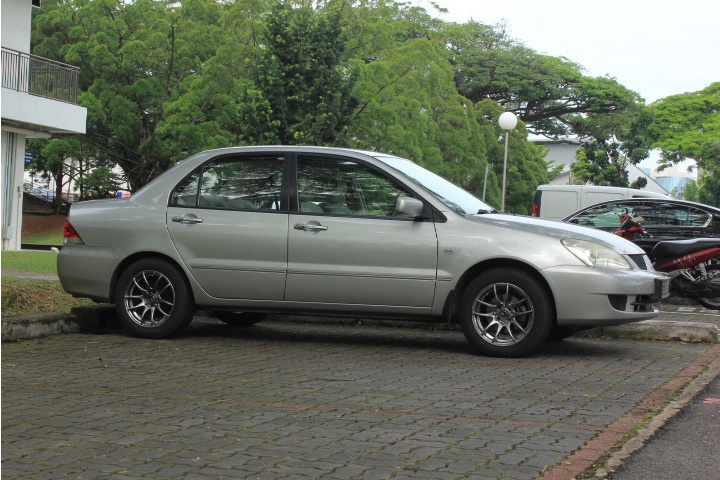}}
	\subfigure{
	\includegraphics[width=\WIDTHSIX]{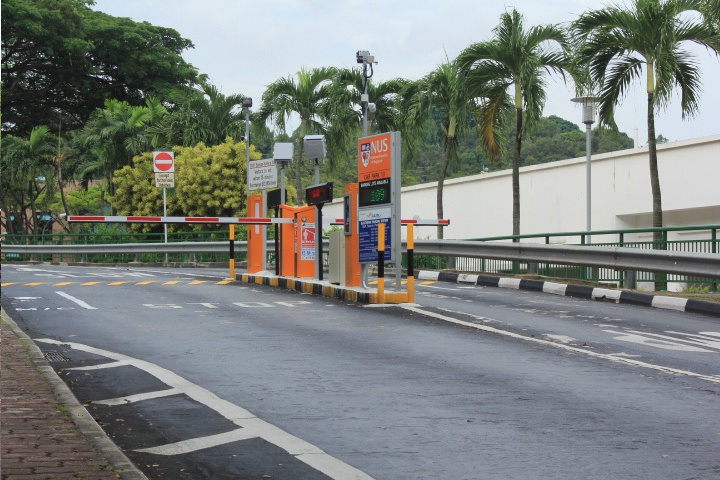}}
	\subfigure{
	\includegraphics[width=\WIDTHSIX]{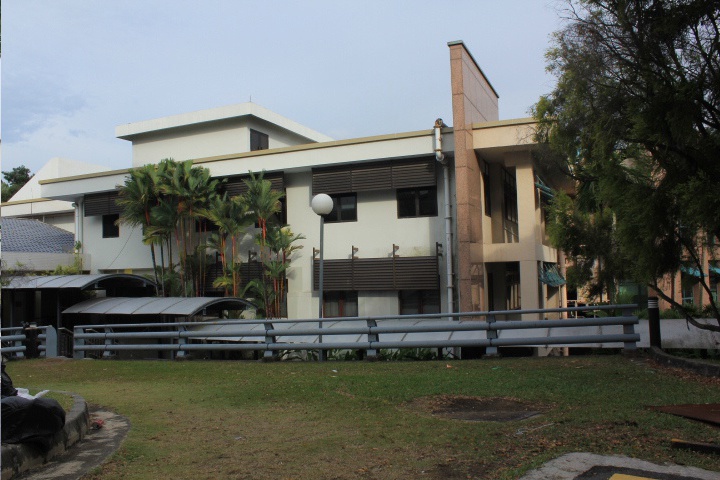}}
\caption{Samples of our dataset. Top: The images degraded with raindrops. Bottom: The corresponding ground-truth images.}
\label{fig:dataset_sample}
\end{figure}

In total, we captured 1119 pairs of images, with various background scenes and raindrops. We used Sony A6000 and Canon EOS 60 for the image acquisition. Our glass slabs have the thickness of 3 mm and attached to the camera lens. We set the distance between
the glass and the camera varying from 2 to 5 cm to
generate diverse raindrop images, and to minimize the reflection effect of the glass. Fig.~\ref{fig:dataset_sample} shows some samples of our data. 

%-------------------------------------------------------------------------

%-------------------------------------------------------------------------
\section{Experimental Results}
\label{sec:results}

\paragraph{Quantitative Evaluation.}

Table 1 shows the quantitative comparisons between our method and other existing methods: Eigen13 \cite{eigen2013restoring}, Pix2Pix \cite{isola2016image}. As shown in the table, compared to these two,  our PSNR and SSIM values are higher. This indicates that our method can generate results  more similar to the groundtruths.

We also compare our whole attentive GAN with some parts of our own network: A (autoencoder alone without the attention map), A+D (non-attentive autoencoder plus non-attentive discriminator), A+AD (non-attentive autoencoder plus attentive discriminator). Our whole attentive GAN is indicated by AA+AD (attentive autoencoder plus attentive discriminator). As shown in the evaluation table, AA+AD  performs better than the other possible configurations. This is the quantitative evidence that the attentive map is needed by both the generative and discriminative networks.

\begin{table}
	\normalsize
	\centering
	\renewcommand\arraystretch{1.1}
	\label{tab:quantitative_eval}
	\begin{tabular}{l|c|c}
		\hline
		\multirow{2}{*}{Method} &
		\multicolumn{2}{c}{Metric} \\
		\cline{2-3}
		& PSNR & SSIM  \\
		\hline
		Eigen13~\cite{eigen2013restoring} & 28.59 & 0.6726\\
		\hline
		Pix2Pix~\cite{isola2016image} & 30.14 & 0.8299\\
		\hline
		 A & 29.25 & 0.7853\\
		\hline
		 A + D & 30.88 & 0.8670\\
		\hline
		 A + AD & 30.60 & 0.8710\\
		\hline
		Ours (AA+AD) & \textbf{31.57} & \textbf{0.9023}\\
		\hline
	\end{tabular}
	
	\vspace{0.5cm}
	
	\caption{Quantitative evaluation results. A is our contextual autoencoder alone. A+D  is autoencoder plus discriminator. A+AD is autoencoder plus attentive discriminator. AA+ AD is our complete architecture: Attentive autoencoder plus attentive discriminator.}
\end{table}

\paragraph{Qualitative Evaluation.}

\begin{figure*}
\centering
\subfigure{
\includegraphics[width=\WIDTHFIVE]{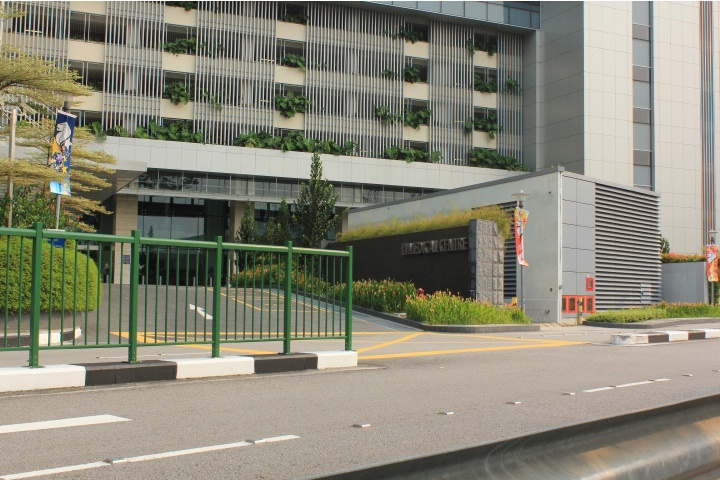}}
\subfigure{
\includegraphics[width=\WIDTHFIVE]{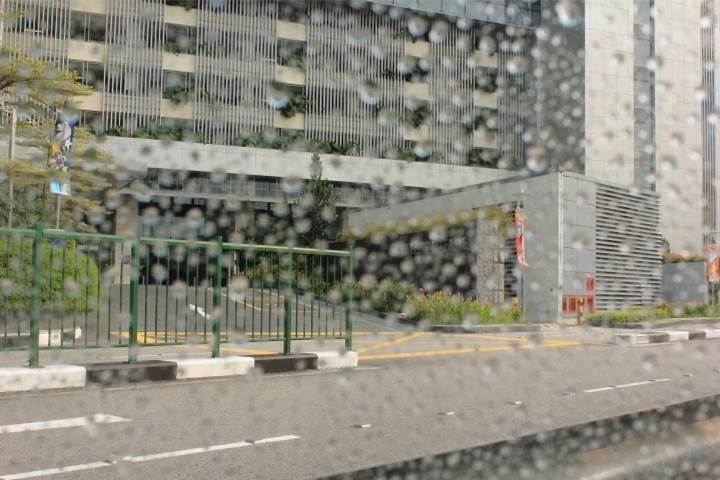}}
\subfigure{
\includegraphics[width=\WIDTHFIVE]{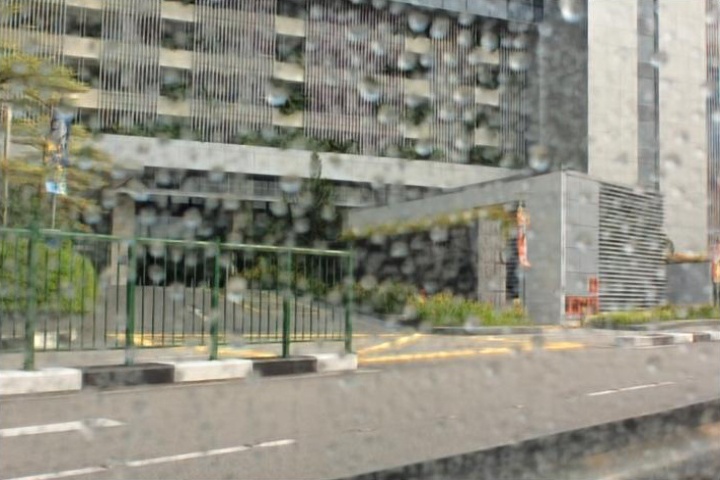}}
\subfigure{
\includegraphics[width=\WIDTHFIVE]{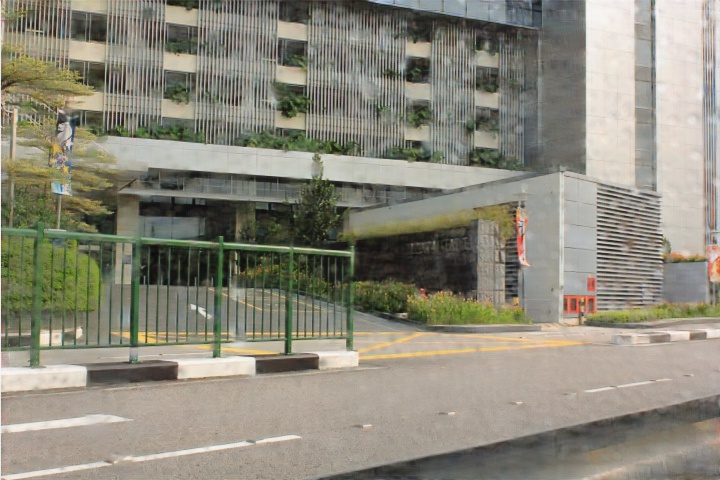}}
\subfigure{
\includegraphics[width=\WIDTHFIVE]{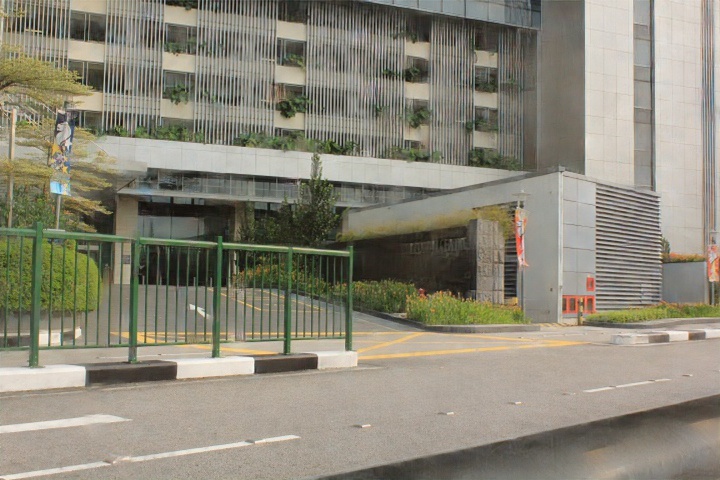}}\vspace{-2.5mm}
\subfigure{
\includegraphics[width=\WIDTHFIVE]{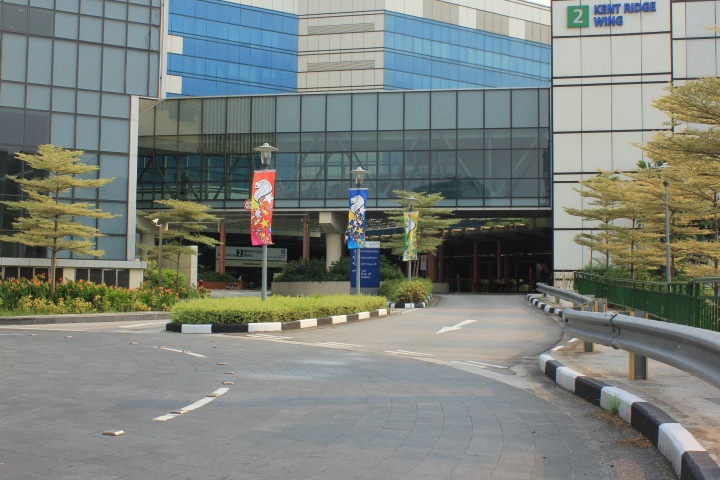}}
\subfigure{
\includegraphics[width=\WIDTHFIVE]{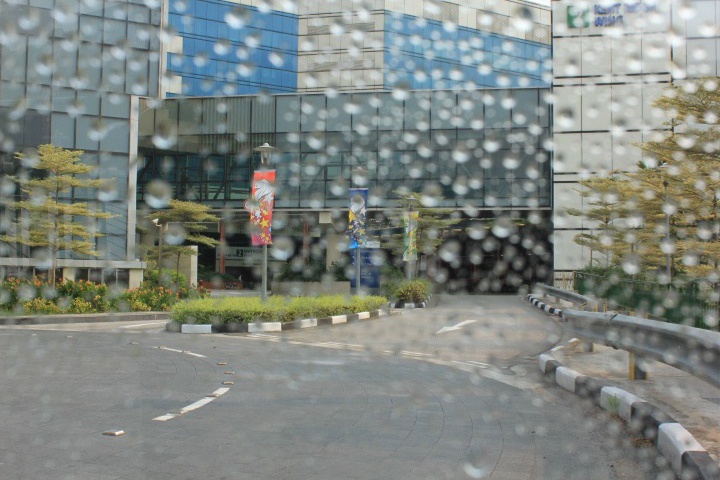}}
\subfigure{
\includegraphics[width=\WIDTHFIVE]{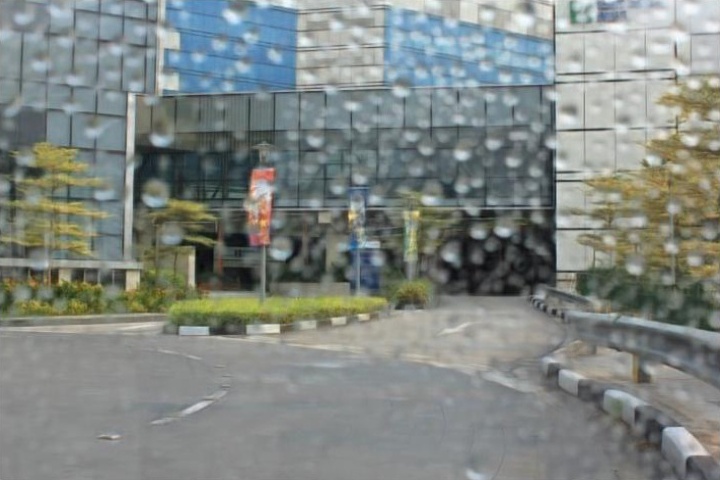}}
\subfigure{
\includegraphics[width=\WIDTHFIVE]{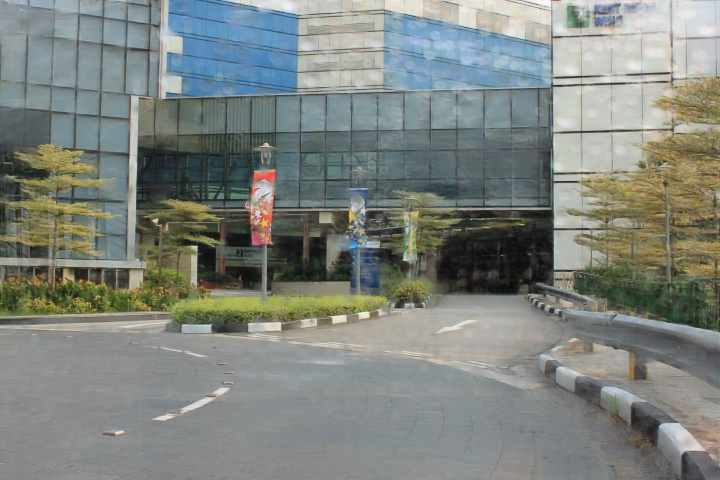}}
\subfigure{
\includegraphics[width=\WIDTHFIVE]{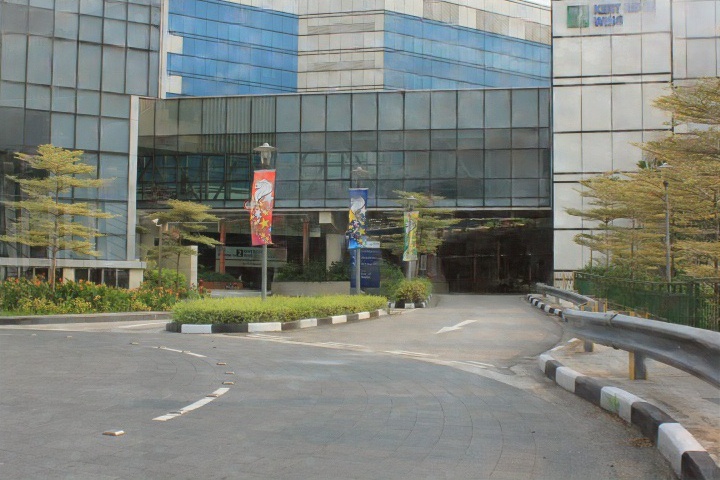}}\vspace{-2.5mm}
\subfigure{
\includegraphics[width=\WIDTHFIVE]{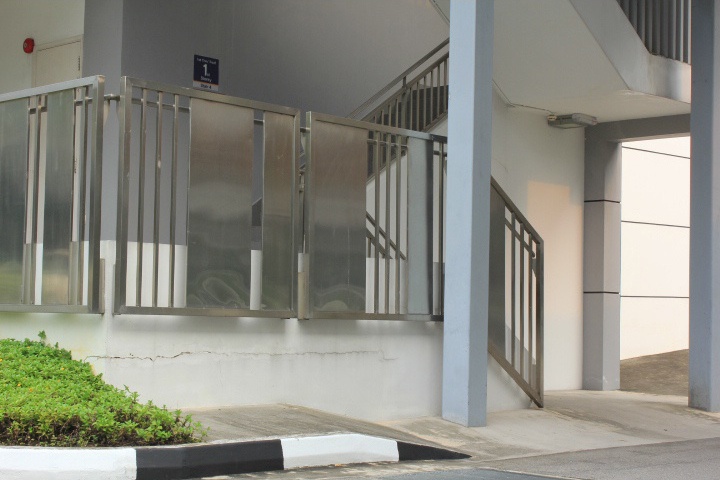}}
\subfigure{
\includegraphics[width=\WIDTHFIVE]{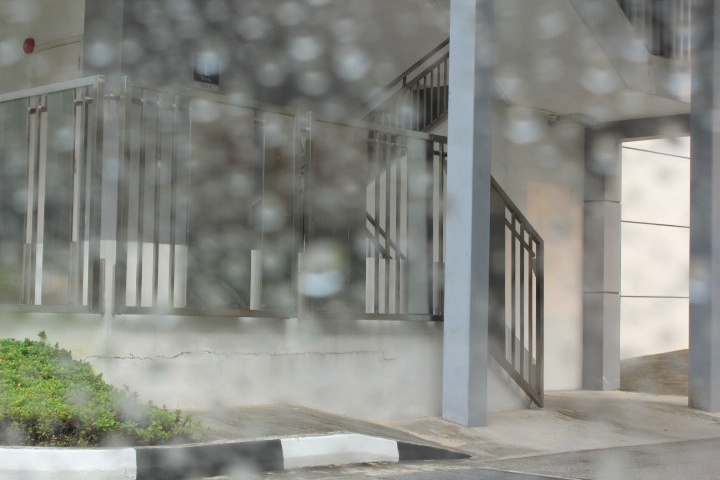}}
\subfigure{
\includegraphics[width=\WIDTHFIVE]{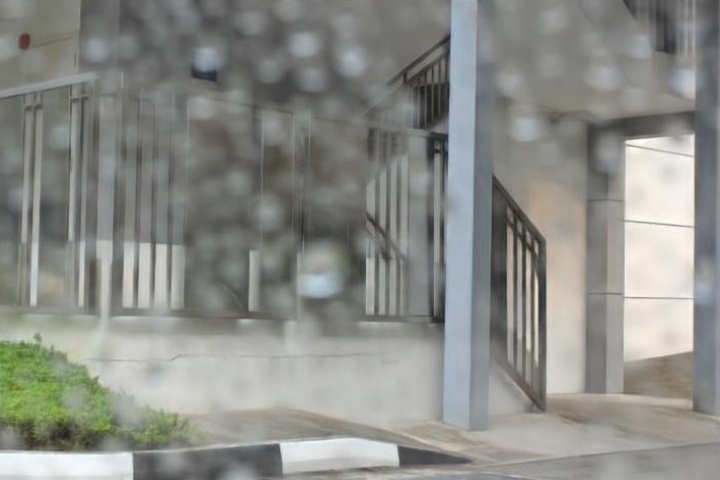}}
\subfigure{
\includegraphics[width=\WIDTHFIVE]{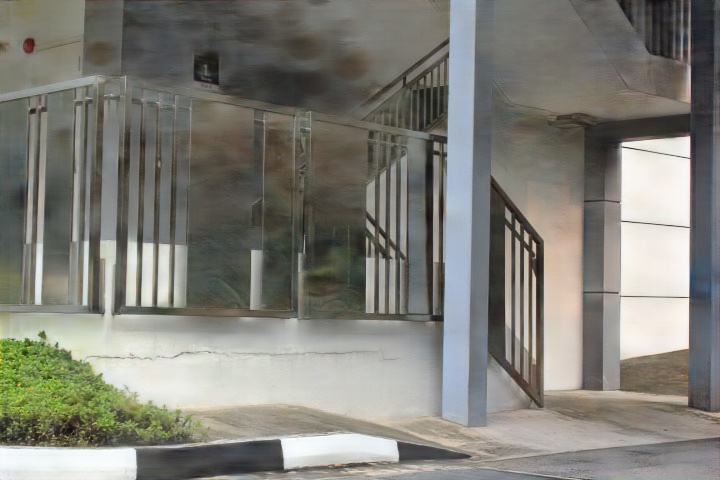}}
\subfigure{
\includegraphics[width=\WIDTHFIVE]{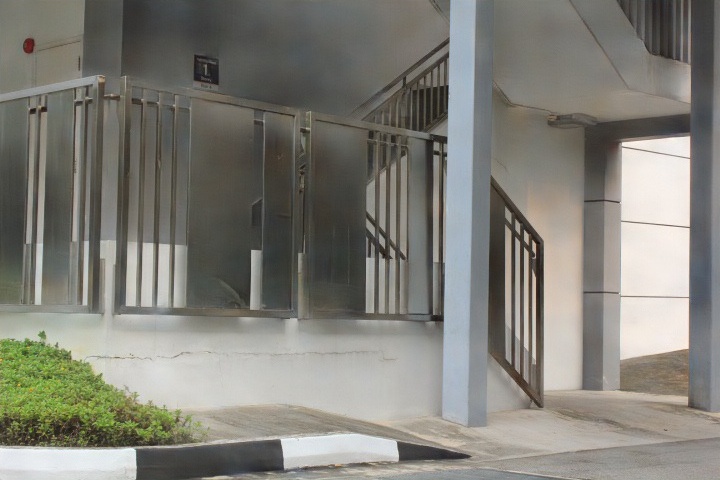}}\vspace{-2.5mm}\addtocounter{subfigure}{-15}
\subfigure[Ground truth]{
\includegraphics[width=\WIDTHFIVE]{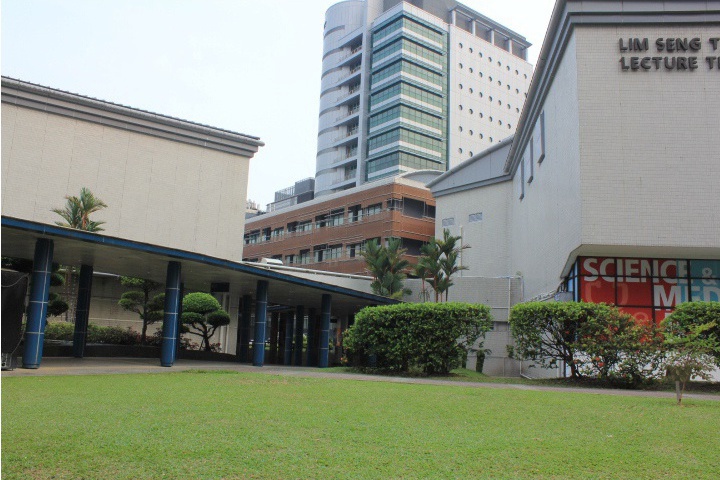}}
\subfigure[Raindrop image]{
\includegraphics[width=\WIDTHFIVE]{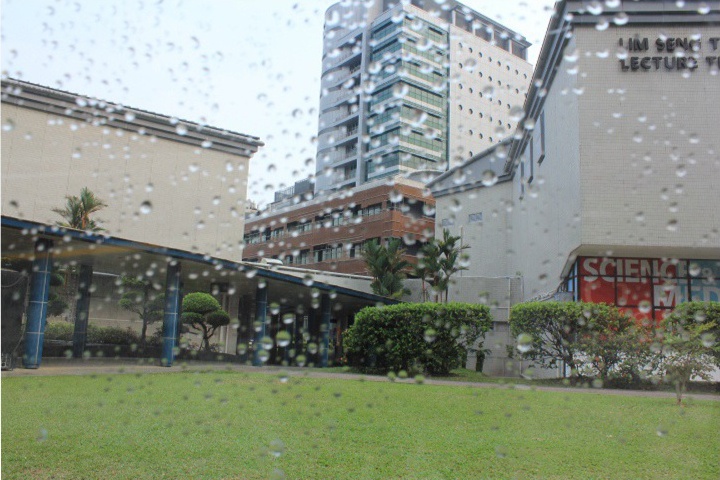}}
\subfigure[Eigen\protect\cite{eigen2013restoring}]{
\includegraphics[width=\WIDTHFIVE]{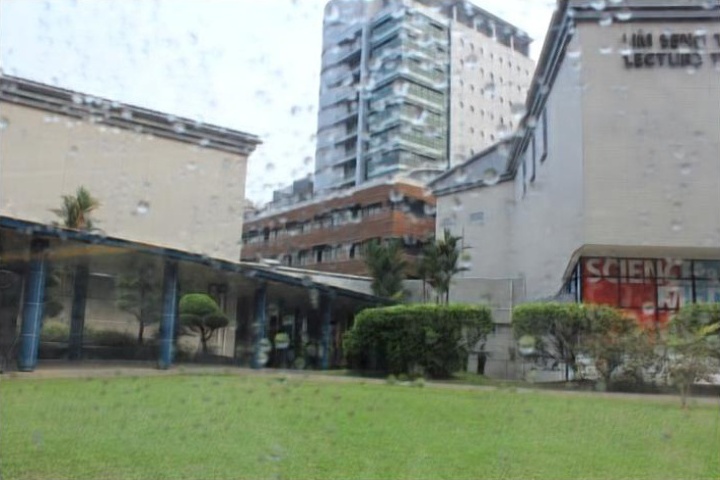}}
\subfigure[Pix2pix-cGAN\protect\cite{isola2016image}]{
\includegraphics[width=\WIDTHFIVE]{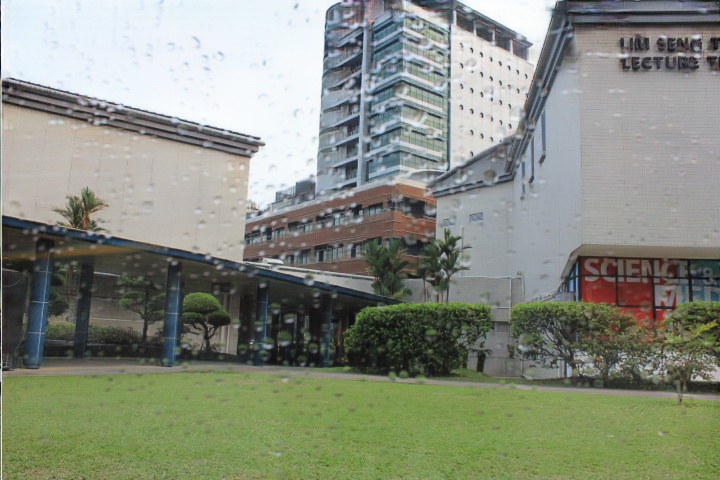}}
\subfigure[Our method]{
\includegraphics[width=\WIDTHFIVE]{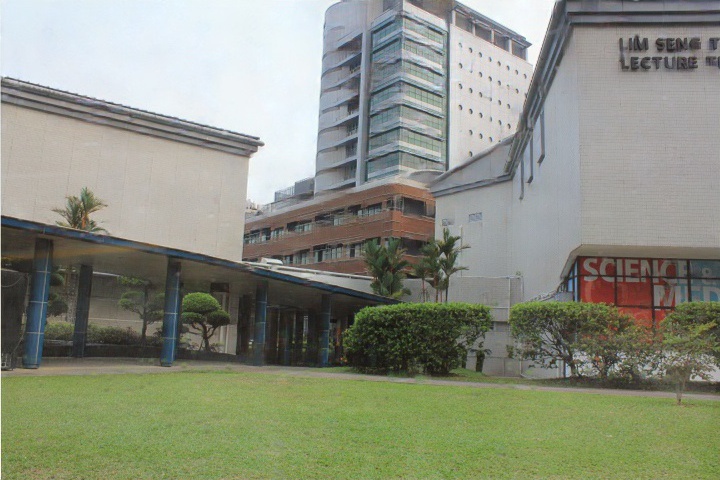}}
\caption{Results of comparing a few different methods. From left to right:
ground truth, raindrop image (input), Eigen13 \cite{eigen2013restoring}, Pix2Pix \cite{isola2016image} and our method. Nearly all raindrops are removed by our method despite the diversity of their colors, shapes and transparency.}
\label{fig:compare1}
\end{figure*}

\begin{figure*}
	\centering
	\subfigure{
	\includegraphics[width=\WIDTHFIVE]{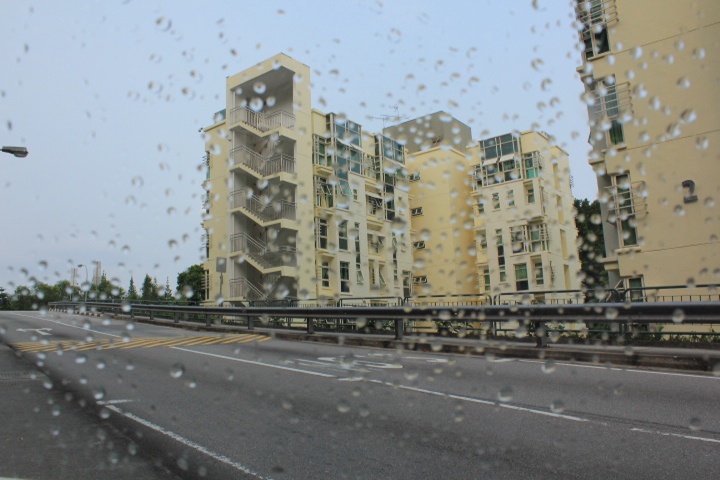}}
	\subfigure{
	\includegraphics[width=\WIDTHFIVE]{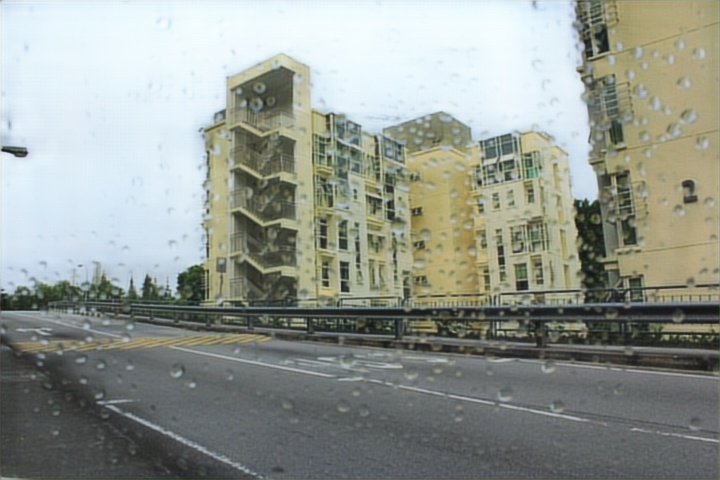}}
	\subfigure{
	\includegraphics[width=\WIDTHFIVE]{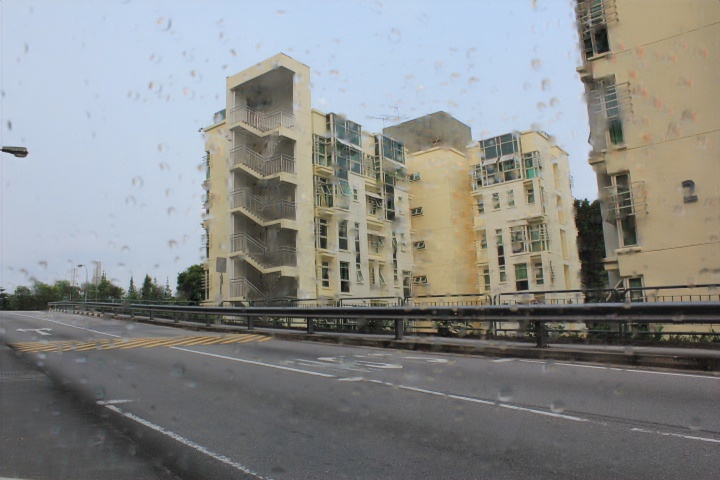}}
	\subfigure{
	\includegraphics[width=\WIDTHFIVE]{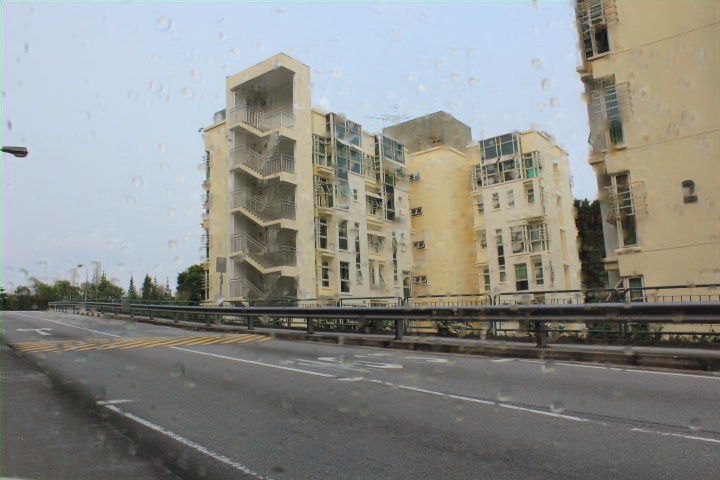}}
	\subfigure{
	\includegraphics[width=\WIDTHFIVE]{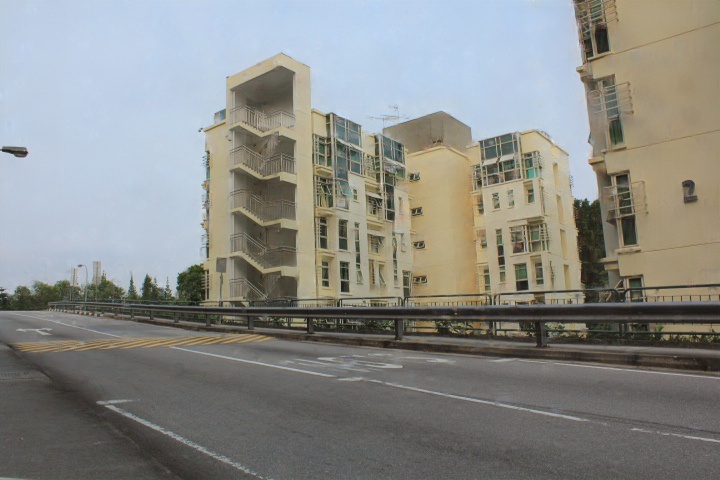}}\vspace{-2.5mm}\addtocounter{subfigure}{-5}
	\subfigure[Input]{
	\includegraphics[width=\WIDTHFIVE]{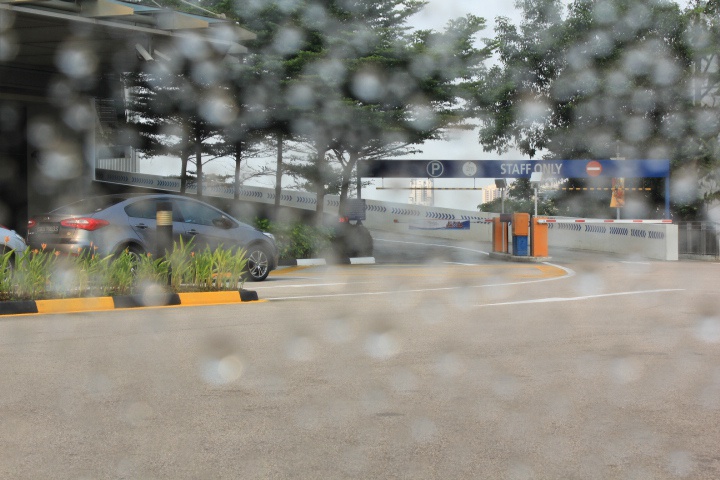}}
	\subfigure[A]{
	\includegraphics[width=\WIDTHFIVE]{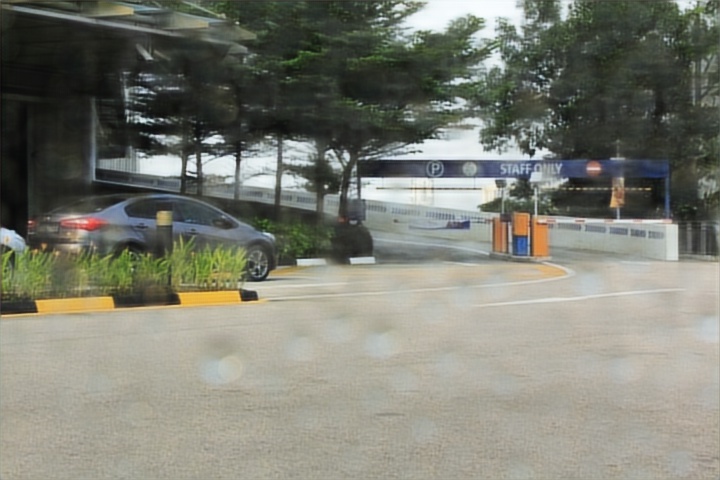}}
	\subfigure[A+D]{
	\includegraphics[width=\WIDTHFIVE]{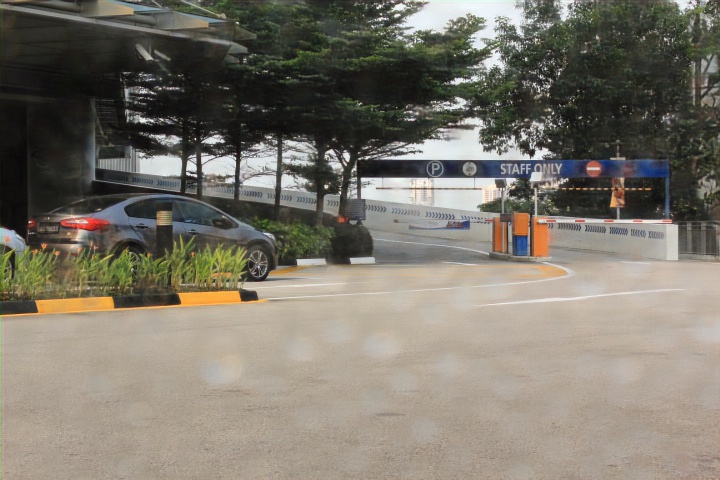}}
	\subfigure[A+AD]{
	\includegraphics[width=\WIDTHFIVE]{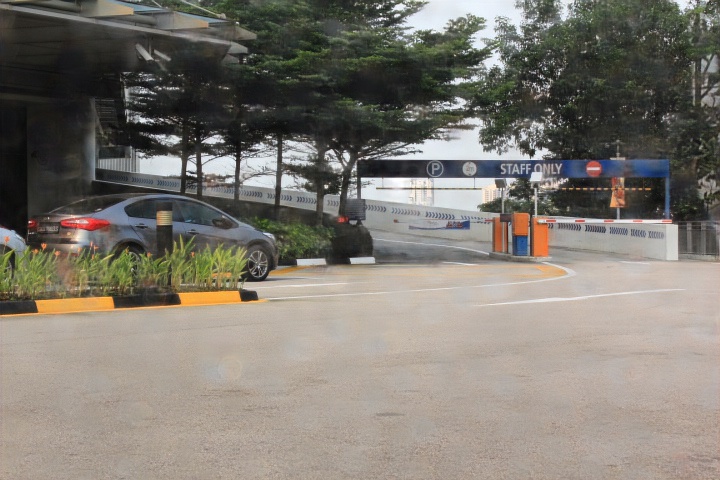}}
	\subfigure[AA+AD]{
	\includegraphics[width=\WIDTHFIVE]{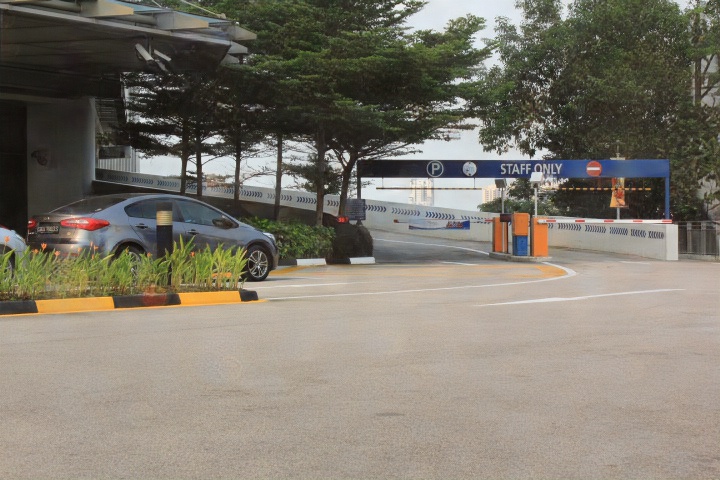}}
	\caption{Comparing  some parts of our network architecture. From left to right: Input, A, A+D, A+AD, our complete architecture (AA+AD). %The result demonstrates that: (1) the discriminator could help obtain the style information of the input image(see(b)(c)), (2) attentive discriminator has better denoising ability as well as improves global and local consistency(see(c)(d)), (3) our network architecture has better performance in removing nearly all raindrops. Overall, every part of our network architecture is useful and the combination reaches the peak performance.% 
	}
	\label{fig:ablation}
\end{figure*}

\begin{figure}
	\centering
	\centering
	\subfigure{
	\includegraphics[width=\WIDTHTHREE]{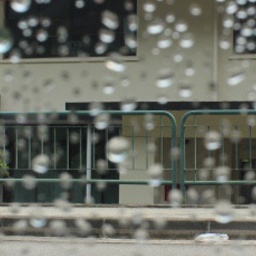}}
	\subfigure{
	\includegraphics[width=\WIDTHTHREE]{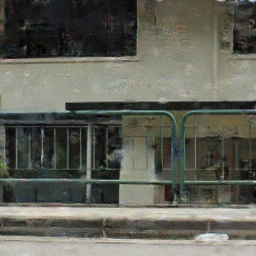}}
	\subfigure{
	\includegraphics[width=\WIDTHTHREE]{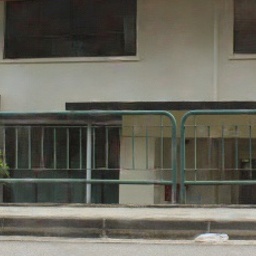}}\vfill\vspace{-2.5mm}\addtocounter{subfigure}{-3}
	\subfigure[Input]{
	\includegraphics[width=\WIDTHTHREE]{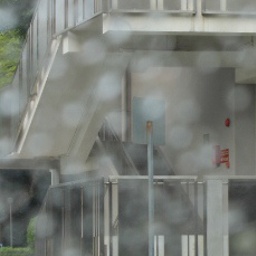}}
	\subfigure[Pix2pix-cGAN]{
	\includegraphics[width=\WIDTHTHREE]{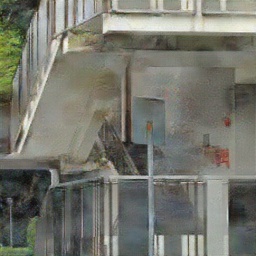}}
	\subfigure[Our method]{
	\includegraphics[width=\WIDTHTHREE]{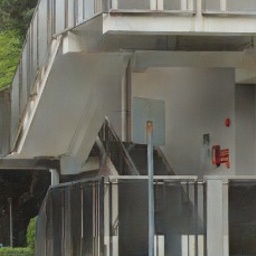}}
	\caption{A closer look at the comparison between our outputs and Pix2Pix's outputs. Our outputs have less artifacts and better restored structures.}
	\label{fig:compare2}
\end{figure}

Fig.~\ref{fig:compare1} shows the results of Eigen13 \cite{eigen2013restoring} and Pix2Pix \cite{isola2016image} in comparison to our results. As can be seen, our method is considerably more effective in removing raindrops compared to Eigen13 and Pix2Pix. In Fig.~\ref{fig:ablation}, we also compare our whole network (AA+AD) with other possible configurations from our architectures (A, A+D, A+AD). Although A+D is qualitatively better than A, and A+AD is better than A+D, our overall network is more effective than A+AD. This is the qualitative evidence that, again, the attentive map is needed by both the generative and discriminative networks.

Fig.~\ref{fig:compare2} provides another comparison between our results and Pix2Pix's results. As can be observed, our outputs have less artifacts and have better restored structures.

%Note that, since Pix2Pix \cite{isola2016image} requires input and output images to be in the size of 256$\times$256, we downscale our original image and up-scale it back using the bilinear interpolation. To make the evaluation more fair as well as comparing detail information, we randomly crop 256$\times$256 images from the test dataset and compare our method with pix2pix-cGAN in Fig.\ref{fig_detail}.

\begin{figure}
\centering
\subfigure[Recognizing result of original image]{
\label{Fig.sub.1_g1}
\includegraphics[width=0.45\textwidth]{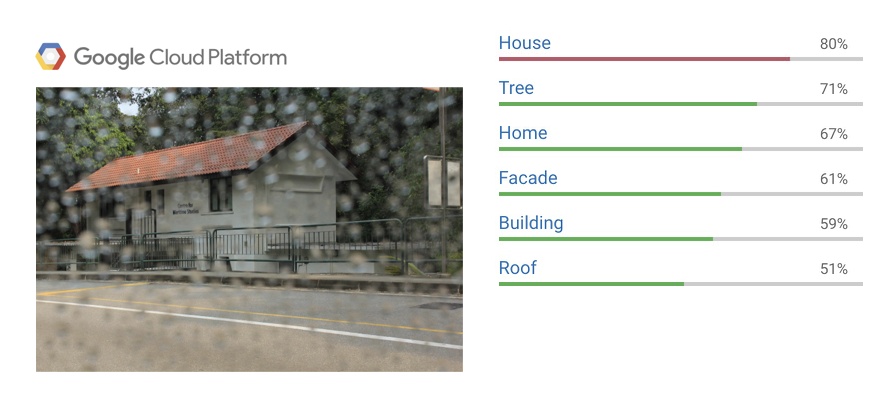}}\vspace{3mm}
\subfigure[Recognizing result of our removal result]{
\label{Fig.sub.2_g1}
\includegraphics[width=0.45\textwidth]{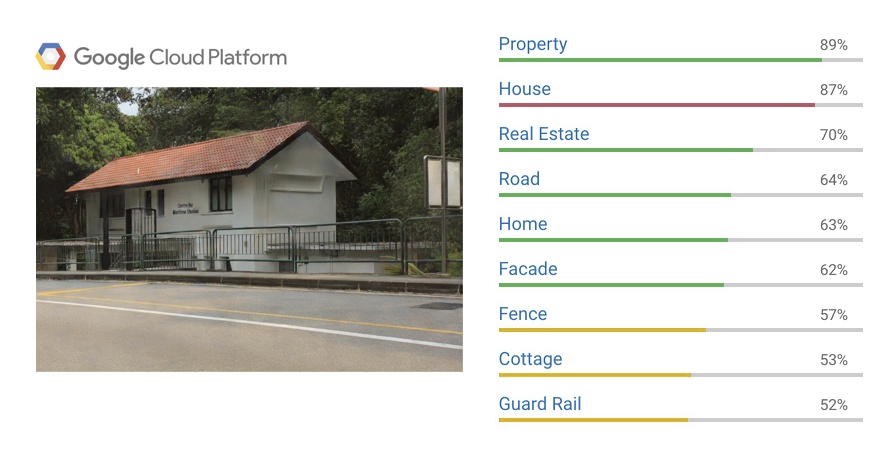}}
\caption{A sample of improving the result of Google Vision API. Our method increases the scores of main object detection as well as the number of the objects recognized.}
\label{fig:app1}
\end{figure}

\begin{figure}
\centering
\subfigure[Score]{
\label{Fig.sub.1_v1}
\includegraphics[width=0.22\textwidth]{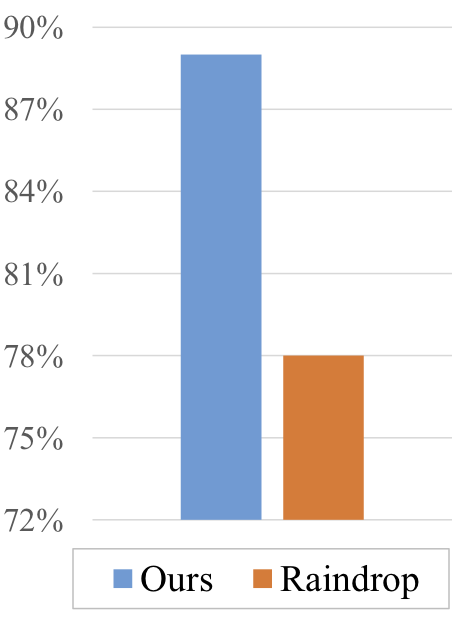}}
\subfigure[Number]{
\label{Fig.sub.2_v2}
\includegraphics[width=0.22\textwidth]{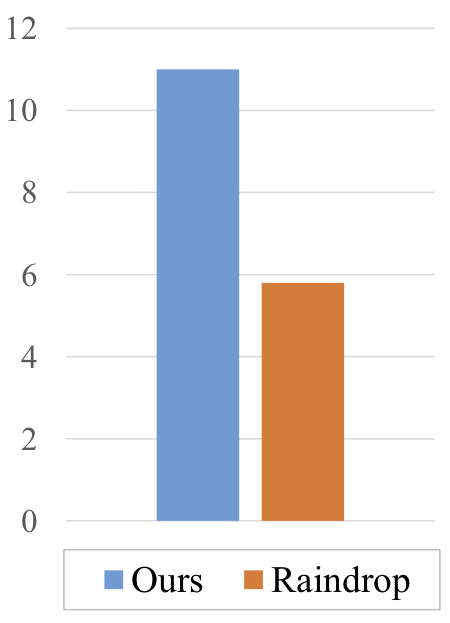}}
\caption{Summary of improvement based on Google Vision API: (a) the average score of identifying the main object in the input image. (b)  the number of object labels recognized. Our method improves the recognization score by 10$\%$ and benefit the recall by 100$\%$ extra object identification.} 
\label{fig:app2}
\end{figure}

\paragraph{Application.}
To provide further evidence that our visibility enhancement could be useful for computer vision applications, we employ Google Vision API (https://cloud.google.com/vision/) to test whether using our outputs can improve the recognition performance.  The results  are shown in Fig.~\ref{fig:app1}. As can be seen, using our output, the general recognition is better than without our visibility enhancement process. Furthermore, we perform evaluation on our test dataset, and Fig.~\ref{fig:app2} shows statistically that using our visibility enhancement outputs significantly outperform those without visibility enhancement, both in terms of the average score of identifying the main object in the input image, and the number of object labels recognized.

\vspace{5mm}

%-------------------------------------------------------------------------
\section{Conclusion}
\label{sec:conclusion}
We have proposed a single-image based raindrop removal method. The method utilizes a generative adversarial network, where the generative network  produces the attention map via an attentive-recurrent network and applies this map along with the input image to generate a raindrop-free image through a contextual autoencoder. Our discriminative network then assesses the validity of the generated output globally and locally. To be able to validate locally, we inject the attention map into the network. Our novelty lies on the use of the attention map in both generative and discriminative network. We also consider that our method is the first method that can handle relatively severe presence of raindrops, which the state of the art methods in raindrop removal fail to handle.

\bibliographystyle{ieee}
\bibliography{egbib}

\end{document}